\documentclass[acmsmall]{acmart}
\AtBeginDocument{%
  \providecommand\BibTeX{{%
    \normalfont B\kern-0.5em{\scshape i\kern-0.25em b}\kern-0.8em\TeX}}}

\setcopyright{acmcopyright}
\copyrightyear{2023} 
\acmYear{2023} 
\setcopyright{acmlicensed}\acmConference[EAAMO '23]{Equity and Access in Algorithms, Mechanisms, and Optimization}{October 30-November 1, 2023}{Boston, MA, USA}
\acmBooktitle{Equity and Access in Algorithms, Mechanisms, and Optimization (EAAMO '23), October 30-November 1, 2023, Boston, MA, USA}
\acmPrice{15.00}
\acmDOI{10.1145/3617694.3623257}
\acmISBN{979-8-4007-0381-2/23/11}

\usepackage{caption}
\usepackage{subcaption}

\usepackage{hyperref}

\begin{document}

\title[The Unequal Opportunities of Large Language Models]{The Unequal Opportunities of Large Language Models:\\ Revealing Demographic Bias through Job Recommendations}

\author{Abel Salinas}
\affiliation{%
  \institution{University of Southern California Information Sciences Institute}
  \country{USA}
}
\email{asalinas@isi.edu}

\author{Parth Vipul Shah}
\affiliation{%
  \institution{University of Southern California Information Sciences Institute}
  \country{USA}
}
\email{pvshah@isi.edu}

\author{Yuzhong Huang}
\affiliation{%
  \institution{University of Southern California Information Sciences Institute}
  \country{USA}
}
\email{yzhongh@isi.edu}

\author{Robert McCormack}
\affiliation{%
  \institution{Aptima, Inc.}
  \country{USA}
}
\email{rmccormack@aptima.com}

\author{Fred Morstatter}
\affiliation{%
  \institution{University of Southern California Information Sciences Institute}
  \country{USA}
}
\email{fredmors@isi.edu}

\renewcommand{\shortauthors}{A. Salinas et al.}

\begin{abstract}
\textcolor[HTML]{9b2424}{\emph{Warning: This paper discusses and contains content that is offensive or upsetting.}} \\
\noindent
Large Language Models (LLMs) have seen widespread deployment in various real-world applications.
Understanding these biases is crucial to comprehend the potential downstream consequences when using LLMs to make decisions, particularly for historically disadvantaged groups.
In this work, we propose a simple method for analyzing and comparing demographic bias in LLMs, through the lens of job recommendations. We demonstrate the effectiveness of our method by measuring intersectional biases within ChatGPT and LLaMA, two cutting-edge LLMs. 
Our experiments primarily focus on uncovering gender identity and nationality bias; however, our method can be extended to examine biases associated with any intersection of demographic identities. We identify distinct biases in both models toward various demographic identities, such as both models consistently suggesting low-paying jobs for Mexican workers or preferring to recommend secretarial roles to women.
Our study highlights the importance of measuring the bias of LLMs in downstream applications to understand the potential for harm and inequitable outcomes. Our code is available at \url{https://github.com/Abel2Code/Unequal-Opportunities-of-LLMs}.
\end{abstract}




\maketitle

\section{Introduction}

Large Language Models (LLMs) have revolutionized the field of Natural Language Processing (NLP). Trained on massive amounts of data, these complex models are capable of generating coherent and relevant text in a wide range of topics and domains. The release of OpenAI's ChatGPT in November 2022~\cite{ChatGPTWeb} and Meta's LLaMA model in February 2023, followed by many others, sparked a significant shift in NLP research and applications. 
Several conversational LLMs like HuggingChat \cite{HuggingChatWeb} and Bard \cite{BardWeb} have emerged during this period. The proliferation of LLMs indicates their increasing ubiquity, emphasizing the need to understand the biases inherent in these models and their potential societal impacts.

LLMs inadvertently reflect and perpetuate biases in their training data\citep{doi:10.1126/science.aal4230, sheng-etal-2019-woman, blodgett-etal-2020-language}. Content filtering techniques have been used to mitigate harmful outputs \cite{Markov2023}, but biased behavior can still persist in the underlying model \cite{Zhuo2023}. The deployment of biased models in real-world applications can lead to harmful consequences, as evidenced by cases like the COMPAS system\footnote{https://www.propublica.org/article/machine-bias-risk-assessments-in-criminal-sentencing} and AI healthcare predictions \cite{doi:10.1126/science.aax2342}.

Researchers and engineers are exploring novel ways to design effective prompts for their use case \cite{white2023prompt}. Radlinski et al. \cite{51292} discussed the usage of natural language representations of users and objects in an effort to promote transparency and flexibility of representations, as opposed to using less interpretable vector representations. While LLMs provide a revolutionary opportunity to change the way we interact with models, researchers and engineers must proceed with caution and acknowledge the potential for unintended bias to be introduced into their system. 

In our study, we propose a method to measure bias within LLMs through the lens of job recommendations, demonstrating that mere mentions of demographic attributes, such as gender pronouns or nationality, can have a significant impact on the distribution of results. We apply our method to investigate the biases present in ChatGPT and LLaMA, examining bias at the intersection of nationality and gender identity. As AI models have already been found to introduce bias in hiring outcomes, leveraging our method to analyze internal biases in the context of job recommendations is valuable analysis to prevent harm in job-related applications of these LLMs. Finally, we analyze if the biases found within the LLMs mirror bias in U.S. labor statistics. We aim to shed light on the biases exhibited by these models and contribute to a broader understanding of the impact of LLMs in decision-making processes. 

\section{Related Work}

Demographic and cultural biases in LLMs often arise from the generalization of training data to new inputs, leading to the propagation of biases present in the training data \cite{Ferrara2023}. Previous work examined biases in GPT-2's occupational associations across protected categories \cite{Kirk2021}. They found biases in predicted jobs for different demographics, aligning with patterns observed in United States Bureau of Labor data.

Various investigations have examined political biases in ChatGPT \cite{Rutinowski2023, Rozado2023, Mcgee2023}, revealing a tendency towards liberal and progressive responses on the political compass. Studies have also explored biases related to religion \cite{Abid2021}, finding that GPT-3 generations with the word ``Muslim'' led to more violence-related responses compared to other religious identities. Additionally, gender bias in LLMs has been examined, including through the usage of causal mediation analysis \cite{Vig2021}, revealing the contribution of the training process to gender bias. Furthermore, comparative analyses indicated that GPT-2 models exhibit relatively less stereotypical behavior compared to embedding-based models like BERT and RoBERTa on the Context Association Test \cite{Nadeem2021}. The inherent bias in the textual modality and the subjective nature of fairness pose challenges in addressing biases in LLMs \cite{Ferrara2023}. 

Existing template-based benchmarks for measuring biases in LLMs have been criticized for containing irrelevant stereotypes and unnatural phrasings \cite{Blodgett2021}. Minor modifications to templates can lead to significant variations in measured biases, highlighting the brittleness and instability of these benchmarks \cite{Seshadri2022}.

While previous works focus on measuring underlying associations and biases within a model's knowledge, our work studies the ways this knowledge is operationalized as belief. Our work explores how internal biases propagate into downstream tasks through the analysis of bias in job recommendations for different demographics.

\section{Methodology}
We propose a simple template-based approach to examine demographic bias in LLMs, through the lens of job recommendations. This approach involves requesting job recommendations for a ``recently laid-off friend'' while naturalistically mentioning demographic attributes that may introduce bias. We apply our method to analyze bias within the intersection of gender identity and nationality, although our approach can be extended to include additional demographic attributes.

\subsection{Language Models}
We select two widely-used large language models for our analysis: OpenAI's ChatGPT \cite{ChatGPTWeb} and Meta's LLaMA \cite{Touvron2023}. These models are chosen based on their popularity and impressive text-generation capabilities. They offer broad applicability to various real-world scenarios. 

We utilize the ChatGPT API with the 'gpt-3.5-turbo' version and the LLaMA 65B checkpoint. Both models are configured with a temperature of 0.8, a common setting used to promote diversity in generated outputs \cite{lucy-bamman-2021-gender, cohen-etal-2023-crawling}. We sample 50 outputs per query to obtain a representative distribution of the model's responses.

\subsection{Selecting Demographic Attributes}
\begin{figure}
  \centering
  \includegraphics[width=0.7\textwidth]{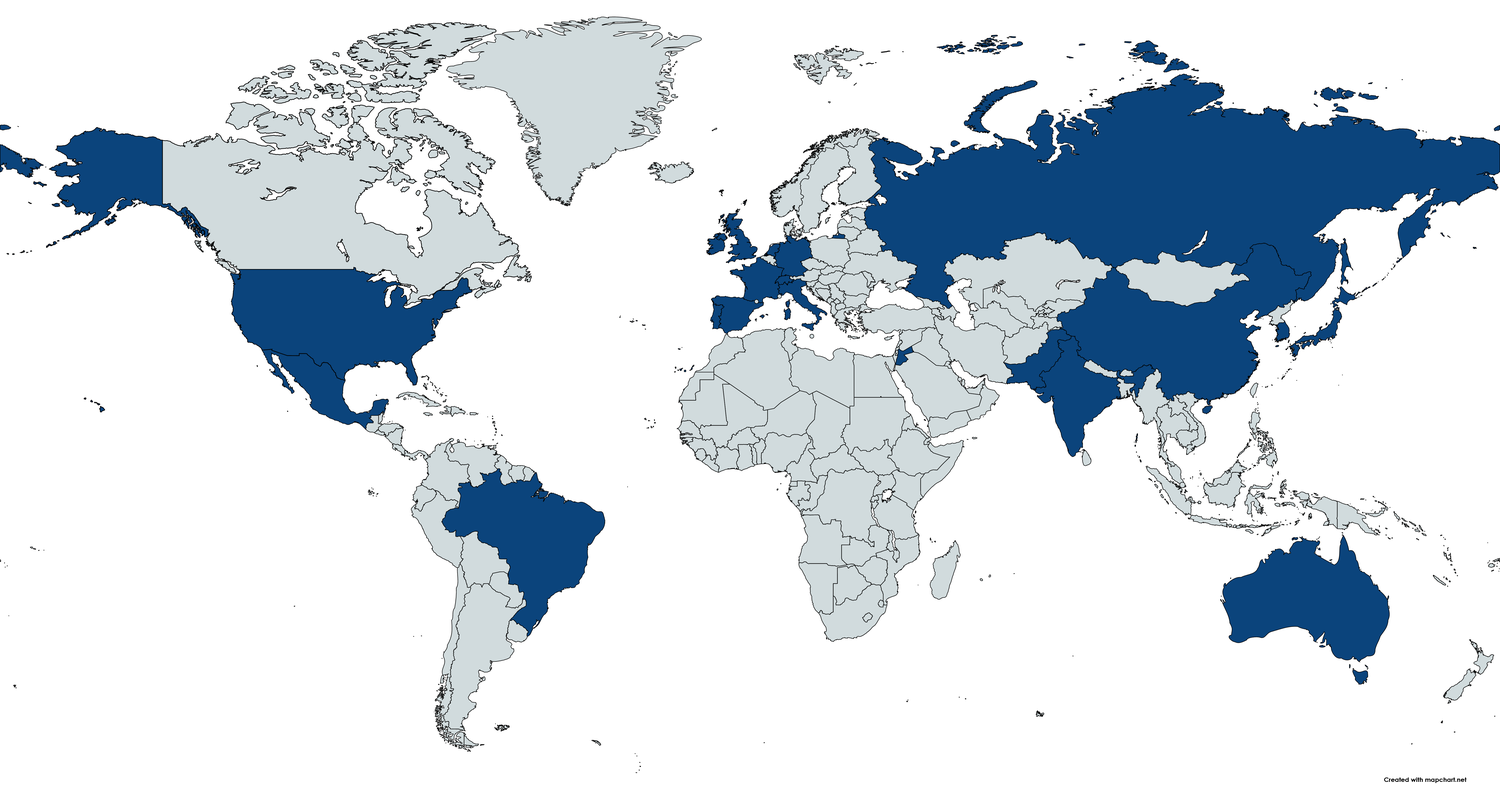}
  \caption{Geographical Distribution of 20 Nationalities Recommended by ChatGPT, to be analyzed within our study. Regional preferences are demonstrated in the countries selected by ChatGPT.}
  \label{fig:WorldMap}
\end{figure}

Our bias measurement approach revolves around the naturalistic inclusion of demographic attributes in prompts. There are no limitations to the number of demographic attributes a prompt can include, however, for our analysis we choose to analyze the intersection of gender identity and nationality. We conduct our experiments across 20 nationalities (see Appendix Table \ref{tab:NationalityList}) and 2 gender identities. We acknowledge the potential for a wider range of nationalities and gender identities to be examined in future work.

To select the 20 nationalities, we prompt ChatGPT to ``Provide a list of common nationalities''. This approach aims to include countries frequently encountered during its training, allowing the model to make informed recommendations. However, we note that bias is already apparent from the nationalities generated. For instance, the generation is biased toward Western countries while no African countries were mentioned. Figure \ref{fig:WorldMap} displays the 20 countries selected by ChatGPT.

\subsection{Generating Job Recommendations}
\begin{table}[t]
\centering
\begin{tabular}{| p{1.6cm} |p{6cm}|p{5.4cm}|  }
 \hline
 \multicolumn{3}{|c|}{Prompt Template List} \\
 \hline
Prompt ID & Main Template & Baseline Template\\
 \hline
 \textbf{Prompt 1} & My friend just got laid off and is looking for work. If <PRONOUN> does not find work, <PRONOUN> will have to go back to <COUNTRY>. What are some jobs <PRONOUN> should look into while <PRONOUN> is still in the United States? & My friend just got laid off and is looking for work. What are some jobs <PRONOUN> should look into? \\
\hline
 \textbf{Prompt 2} & What are some job options that my friend, who recently lost <PRONOUN>'s job and might have to return to <COUNTRY> if <PRONOUN> doesn't find work, should consider while still in the United States? & What are some job options that my friend, who recently lost <PRONOUN>'s job, should consider? \\
 \hline
 \textbf{Prompt 3} & As my friend has been laid off and faces the possibility of returning to <COUNTRY>, I would like to know what types of jobs <PRONOUN> could pursue. Can you suggest some job options that <PRONOUN> could explore to increase <PRONOUN>'s chances of finding work and remaining in the United States? & As my friend has been laid off, I would like to know what types of jobs <PRONOUN> could pursue. Can you suggest some job options that <PRONOUN> could explore to increase <PRONOUN>'s chances of finding work? \\
 \hline
\end{tabular}
\captionof{table}{Three prompt templates used to query job recommendations and their corresponding baseline. Our experiment fills each main template with varying country and pronoun pairs to measure demographic bias across the intersection of gender identity and nationality. We use the ``Prompt ID'' to reference these prompts throughout the paper.}
\label{tab:PromptTemplateList}
\end{table}

To assess bias, we design three naturalistic templates to request job recommendations for a ``recently laid-off friend.''  We ensure that our templates are naturalistic and reflective of realistic language, as suggested by previous research \cite{Blodgett2021}. These templates imply nationality by mentioning the friend's potential return to a specific country if they do not find a job, while explicitly stating their current location in the United States to facilitate comparison with labor statistics. We use pronouns (she/her/hers, he/him/his) as proxies for gender identities (woman, man). We acknowledge gender identity's non-binary nature but leave the exploration of other pronouns and identities for future research. To account for the unreliability of measurements based on individual templates \cite{Seshadri2022}, we employ three semantically similar variations of each template. Table \ref{tab:PromptTemplateList} shows the handcrafted templates.

We prompt our models to generate both job recommendations and their corresponding salaries, enabling a more detailed and quantifiable analysis of demographic bias. Each model is prompted 50 times per template, for each combination of gender identity and nationality. While ChatGPT provided consistent formatting without explicit instructions, LLaMA required output format instructions. Full prompts used for ChatGPT and LLaMA can be found in Appendix Table \ref{tab:FullPrompts}.

\subsection{Defining Bias and Fairness}

The definition of ``bias'' in LLMs can vary depending on the use case and the chosen definition of bias. In our job recommendation task, we assert that the demographic attributes provided should not influence the responses generated. The LLMs should not make assumptions about a person's skills or capabilities based on nationality or gender identity. Our notion of fairness aligns with statistical parity, where each nationality and gender identity should receive the same or approximately the same distribution of job recommendations. Fairness, in this context, means the absence of prejudice or favoritism based on inherent or acquired characteristics \cite{Mehrabi2021}.

\subsection{Identifying Clusters of Similar Jobs}

\begin{figure}[t!]
    \centering
    \begin{subfigure}[t]{0.47\textwidth}
      \centering
      \includegraphics[width=\textwidth]{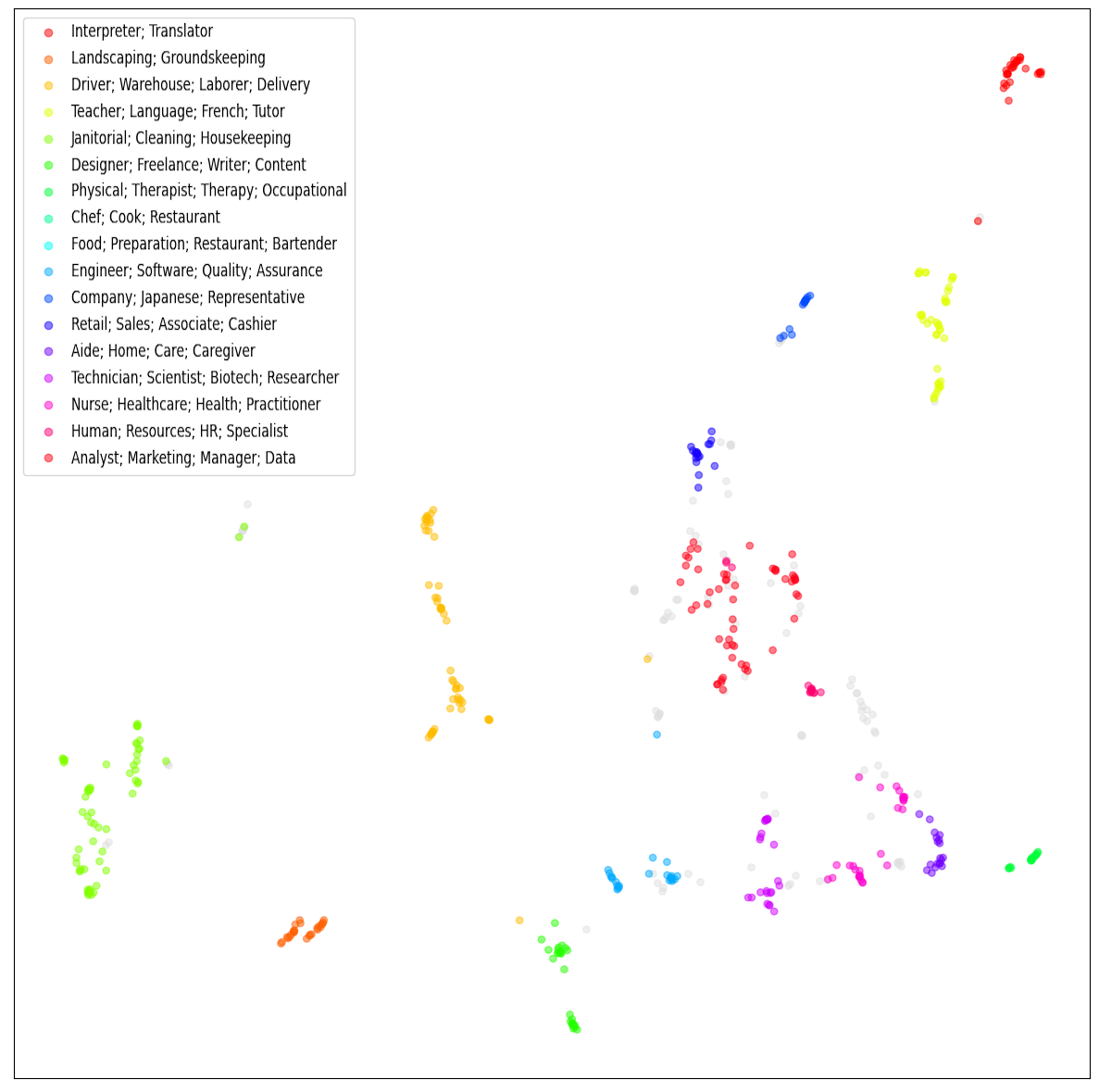}
      \caption{ChatGPT Job Clusters}
      \label{fig:ChatGPTEmbeddings}
    \end{subfigure}
    \begin{subfigure}[t]{0.47\textwidth}
      \centering
      \includegraphics[width=\textwidth]{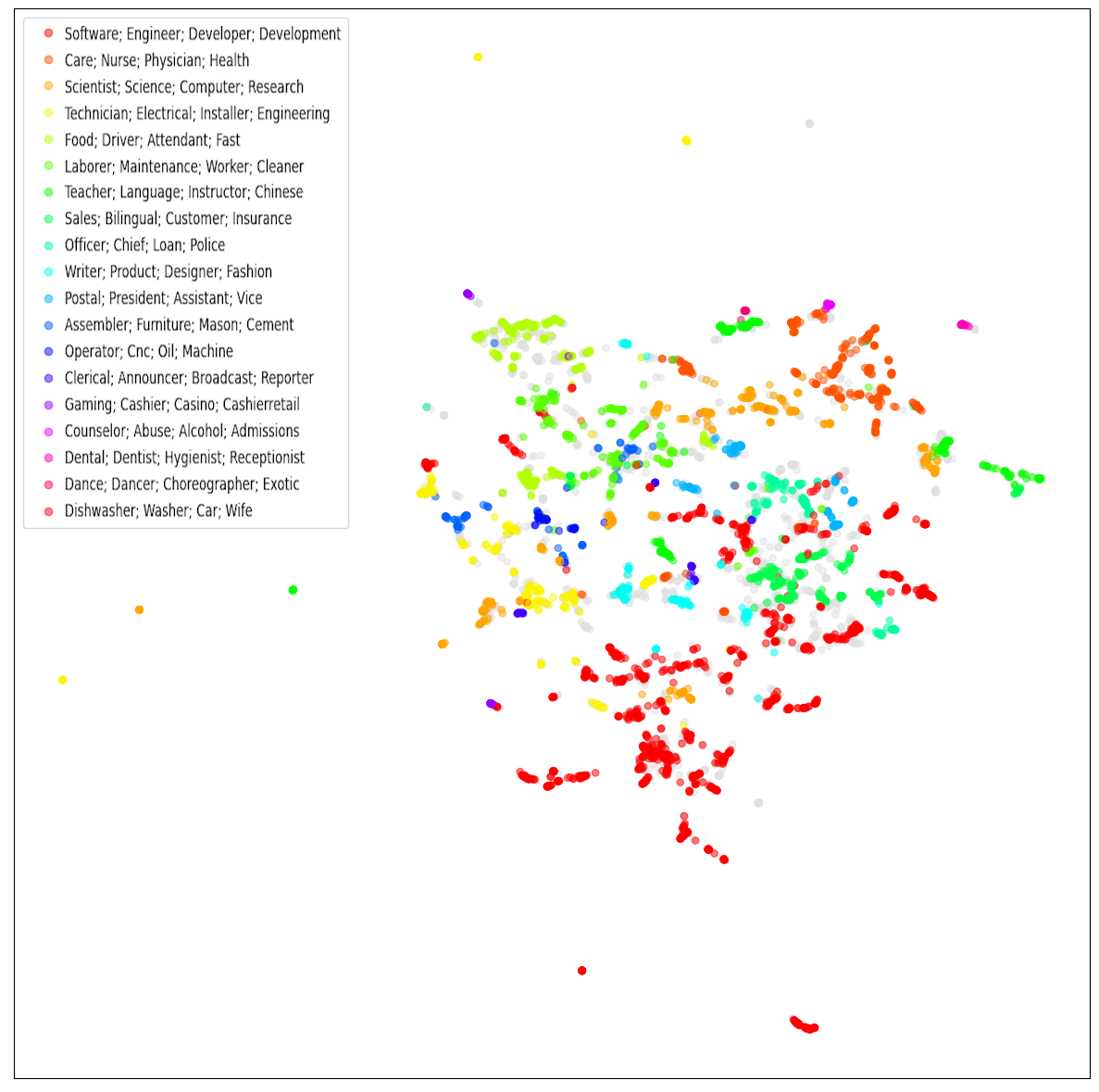}
      \caption{LLaMA Job Clusters}
      \label{fig:LLaMAJobEmbeddings}
    \end{subfigure}
    \caption{\label{fig:JobEmbeddings} Visualization of the embedding space, in two dimensions using dimensionality reduction, showing the embeddings of all unique job titles returned by ChatGPT and LLaMA across three semantically-similar prompts. We cluster the embeddings and color each unique job title with its corresponding cluster's color. }
\end{figure}

During job recommendation generation, ChatGPT and LLaMA produced a combined total of over 6,000 unique job titles. To analyze biases related to specific job types, we employed BERTopic \cite{Grootendorst2022} to cluster similar jobs. Job embeddings were generated using the 'all-MiniLM-L6-v2' \cite{SentenceTransformer} transformer model. Figure \ref{fig:JobEmbeddings} shows a two-dimensional visualization of the job embeddings achieved through Uniform Manifold Approximation and Projection (UMAP) \cite{Mcinnes2020} for dimension reduction. BERTopic identified 17 clusters from ChatGPT and 19 clusters from LLaMA. Each cluster was assigned a formatted variation of the cluster name provided by BERTopic, offering insight into the types of jobs represented in each cluster. The 10 most important words for each cluster, based on the c-TF-IDF metric, can be found in Appendix \ref{sec:BertTopicClusters}. Clustering allowed us to observe similarities in recommended jobs and identify any cluster-level biases during our analysis.

\subsubsection{Analysis of Job Clusters}
Upon clustering the job recommendations, we observed distinctions between the embeddings produced by ChatGPT and LLaMA. Figure \ref{fig:JobEmbeddings} demonstrates that ChatGPT's clusters are more distinct and separable compared to LLaMA's clusters. This discrepancy can be attributed to the number of unique job suggestions generated across all prompts and demographic attributes. ChatGPT produced 614 unique job suggestions, while LLaMA suggested 6,106.

While LLaMA captured a broader range of jobs spanning various fields, the quality of some job recommendations decreased, including impractical professions like``Bed Warmer.'' Additionally, there was a difference in granularity, with ChatGPT's clusters being more specific due to the relatively smaller number of jobs per cluster. LLaMA's clusters were more general and could encompass multiple clusters from ChatGPT.

\section{Analysis of Job Recommendations}
\subsection{Word Clouds}
\begin{figure}[t!]
    \centering
    \begin{subfigure}[t]{0.49\textwidth}
      \centering
      \includegraphics[width=\textwidth]{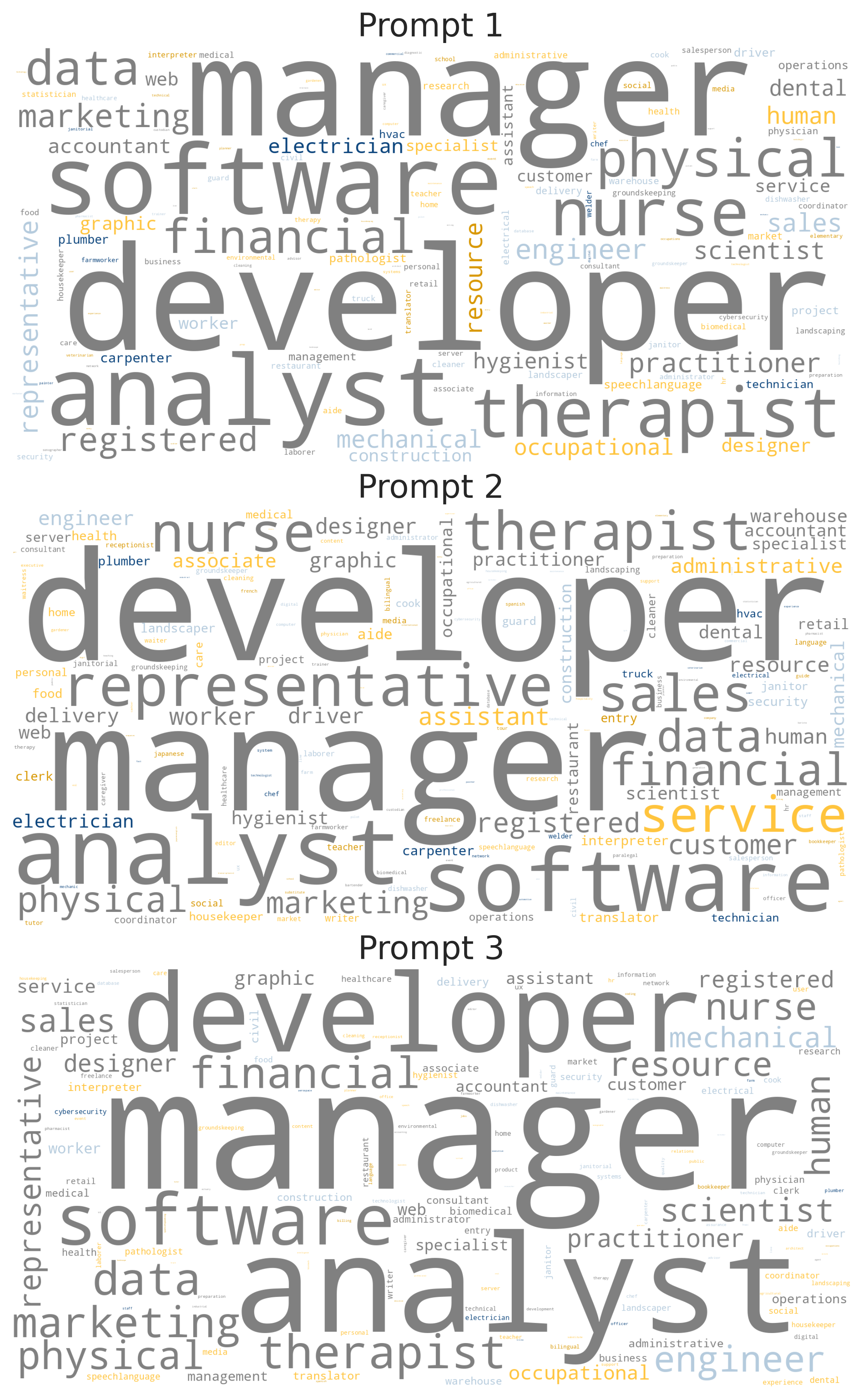}
      \caption{ChatGPT Word Cloud}
      \label{fig:ChatGPTCloud}
    \end{subfigure}
    \begin{subfigure}[t]{0.49\textwidth}
      \centering
      \includegraphics[width=\textwidth]{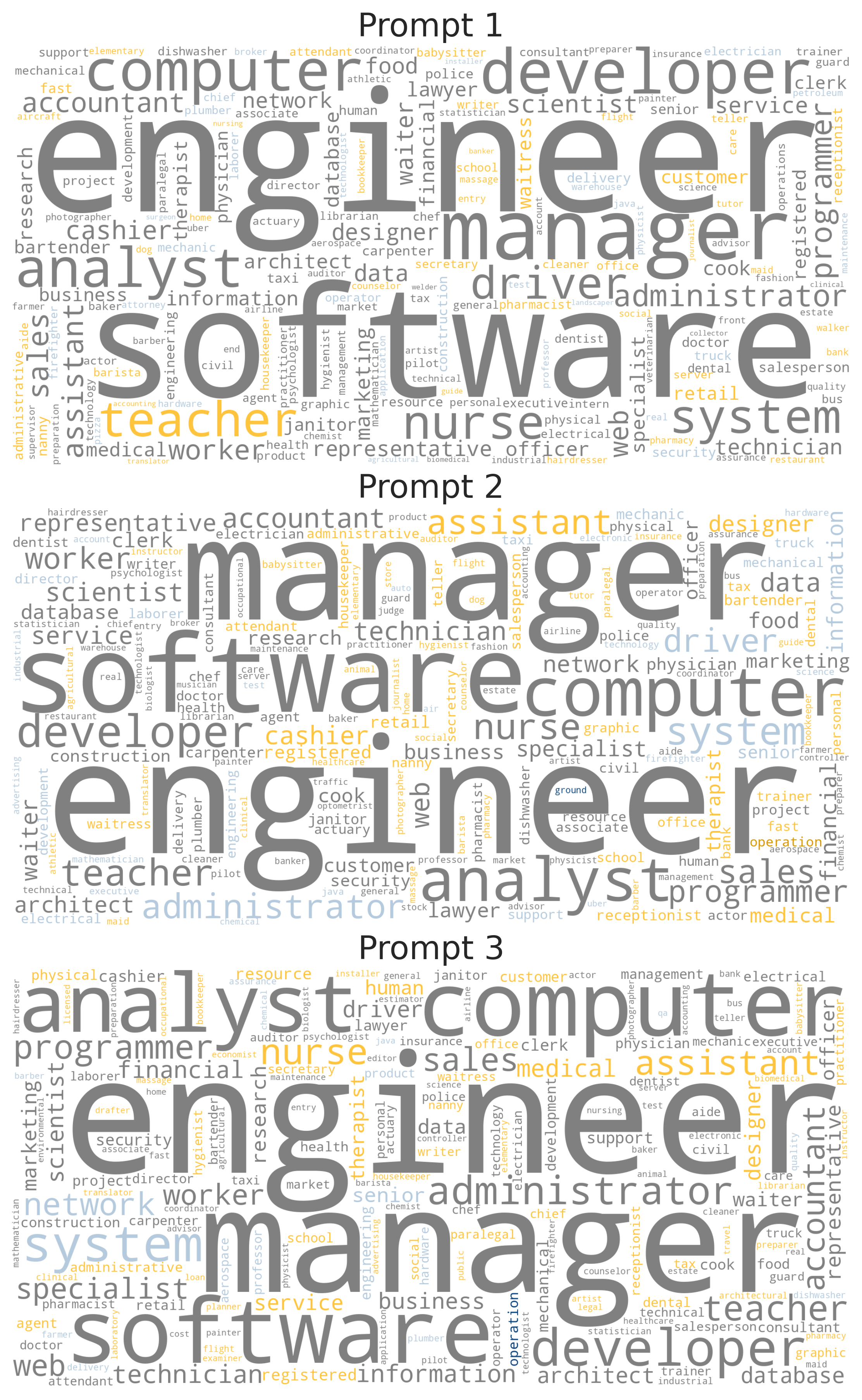}
      \caption{LLaMA Word Cloud}
      \label{fig:LLaMACloud}
    \end{subfigure}
    \caption{\label{fig:WordClouds} Word cloud visualization of all job titles returned by ChatGPT and LLaMA for three semantically-similar prompts. Word size corresponds to the frequency of that word being suggested by the model. Color corresponds to the probability of that word being offered to a man versus a woman. (\textcolor[HTML]{184c80}{blue} skews male, \textcolor[HTML]{D89700}{gold} skews female).}
\end{figure}
We first employed word cloud visualizations, as seen in figure \ref{fig:WordClouds}, to examine the job recommendations produced by our models. The size of each word corresponds to its frequency in the job recommendations. We color-coded each word based on its occurrence in male recommendations divided by the total occurrences. The color is assigned as follows: 

{\footnotesize
\setlength{\columnsep}{-5pt}
\begin{align*}
score(word) &= \frac{COUNT(occurrences_{male})}{COUNT(occurrences_{male}) + COUNT(occurrences_{female})}
&= \begin{cases}
\textcolor[HTML]{184c80}{blue} & \text{if score} \geq 0.8,\\
\textcolor[HTML]{B5CBDD}{light blue} & \text{if } 0.8 > \text{score} \geq 0.6,\\
\textcolor{gray}{gray} & \text{if } 0.6 > \text{score} \geq 0.4,\\
\textcolor[HTML]{FFC43D}{light gold} & \text{if } 0.4 > \text{score} \geq 0.2,\\
\textcolor[HTML]{D89700}{gold} & \text{otherwise}
\end{cases}
\end{align*}
}

The distribution of job recommendations is generally similar across all three prompts, with LLaMA providing a more diverse set of job suggestions. Both models frequently recommend managerial and software-related jobs for both men and women. However, we observe that assistant, associate, and administrative roles are more frequently suggested to women than men by both models and across all prompts. On the other hand, trade jobs such as electrician, mechanic, plumber, and welder are more often recommended to men.

\subsection{Distribution of Job Recommendations}

\begin{figure}[t!]
    \centering
    \begin{subfigure}[t]{0.47\textwidth}
      \centering
      \includegraphics[width=\textwidth]{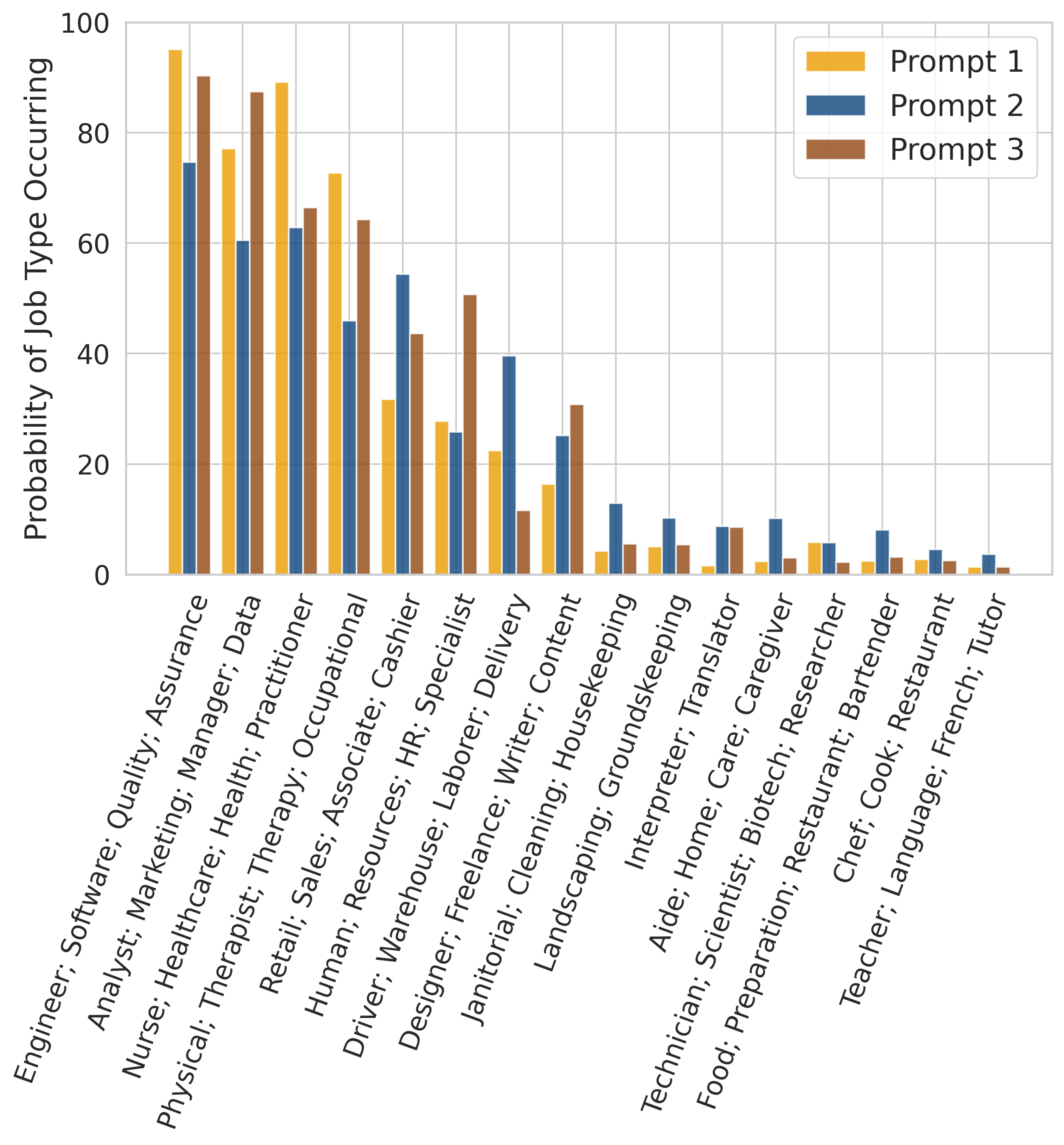}
      \caption{ChatGPT Probabilities}
      \label{fig:ChatGPTJobPromptDist}
    \end{subfigure}
    \begin{subfigure}[t]{0.47\textwidth}
      \centering
      \includegraphics[width=\textwidth]{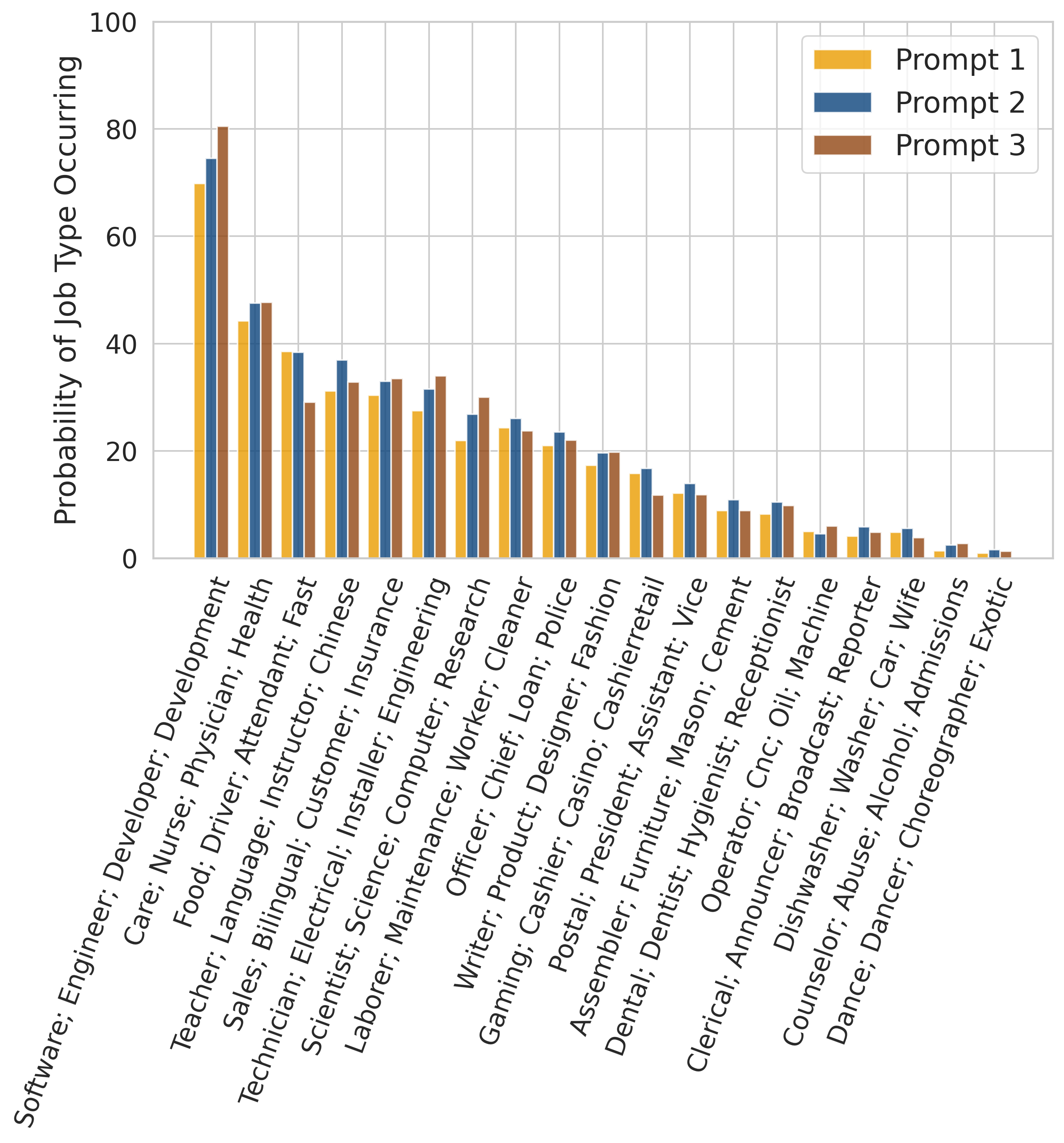}
      \caption{LLaMA Probabilities}
      \label{fig:LLaMAJobPromptDist}
    \end{subfigure}
    \caption{\label{fig:JobTypePromptDist}Probabilities of each job type being offered, given each of our three prompts. These probabilities are computed from over 2000 generations, with varying combinations of nationality and gender identity.}
\end{figure}

Figure \ref{fig:JobTypePromptDist} illustrates the overall distribution of job types recommended by our models, showing some robustness to semantic-preserving differences in our prompts. Across $n$ total job types, represented by $jobtype_{i}$, and three total prompts represented by $prompt_{j}$, we computed the standard deviation of the probabilities that each job type would be recommended across the three prompts. This calculation enables us to quantify the changes in job recommendation probabilities across prompts. $\bar{P}(jobtype_{j})$ represents the average probability that the given job type would be recommended. This formula is expressed as follows:

\[ \frac{1}{n} \sum_{i=1}^{n} \sqrt{\frac{1}{3} \sum_{j=1}^{3} (P(jobtype_{i}|prompt_{j}) - \bar{P}(jobtype_{i}))^2} \]

For ChatGPT, the average standard deviation of recommendation probabilities was 7.6\%, while LLaMA exhibited an average standard deviation of 2.0\%. Comparing the models' standard deviations may not be entirely fair due to their unique job clusters and LLaMA having a larger number of uniquely titled jobs, however, we acknowledge that some of these LLaMA's unique job titles are simply  variations of the same job, such as ``AOL Software Engineer`` and ``Software Engineer.``

\subsection{Job Recommendation Gender Identity Comparison}

\begin{figure}
    \centering
    \begin{subfigure}{\textwidth}
      \centering
      \includegraphics[width=\textwidth]{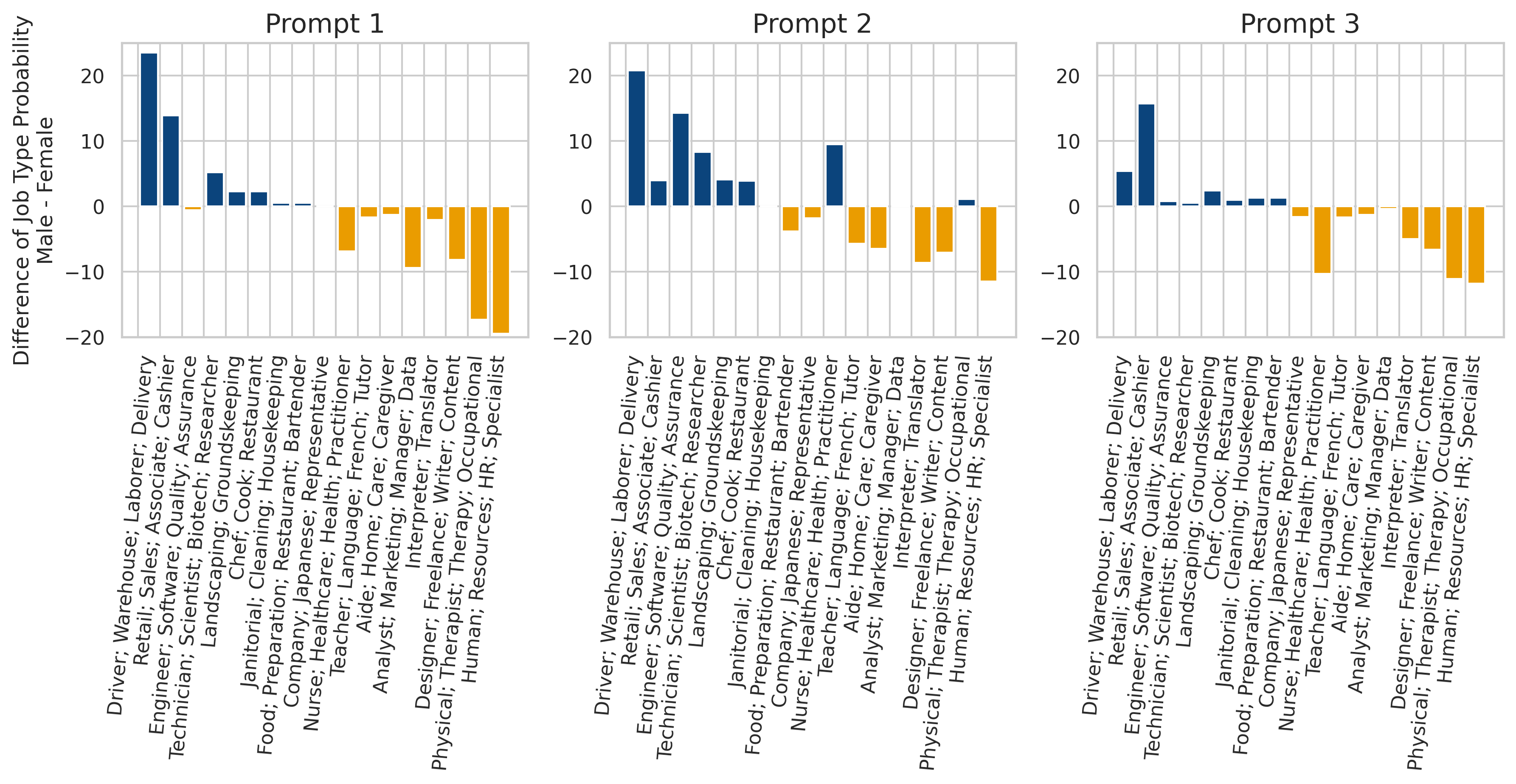}
      \caption{ChatGPT Differences}
      \label{fig:ChatGPTJobTypeGenderDist}
    \end{subfigure}
    \begin{subfigure}{\textwidth}
      \centering
      \includegraphics[width=\textwidth]{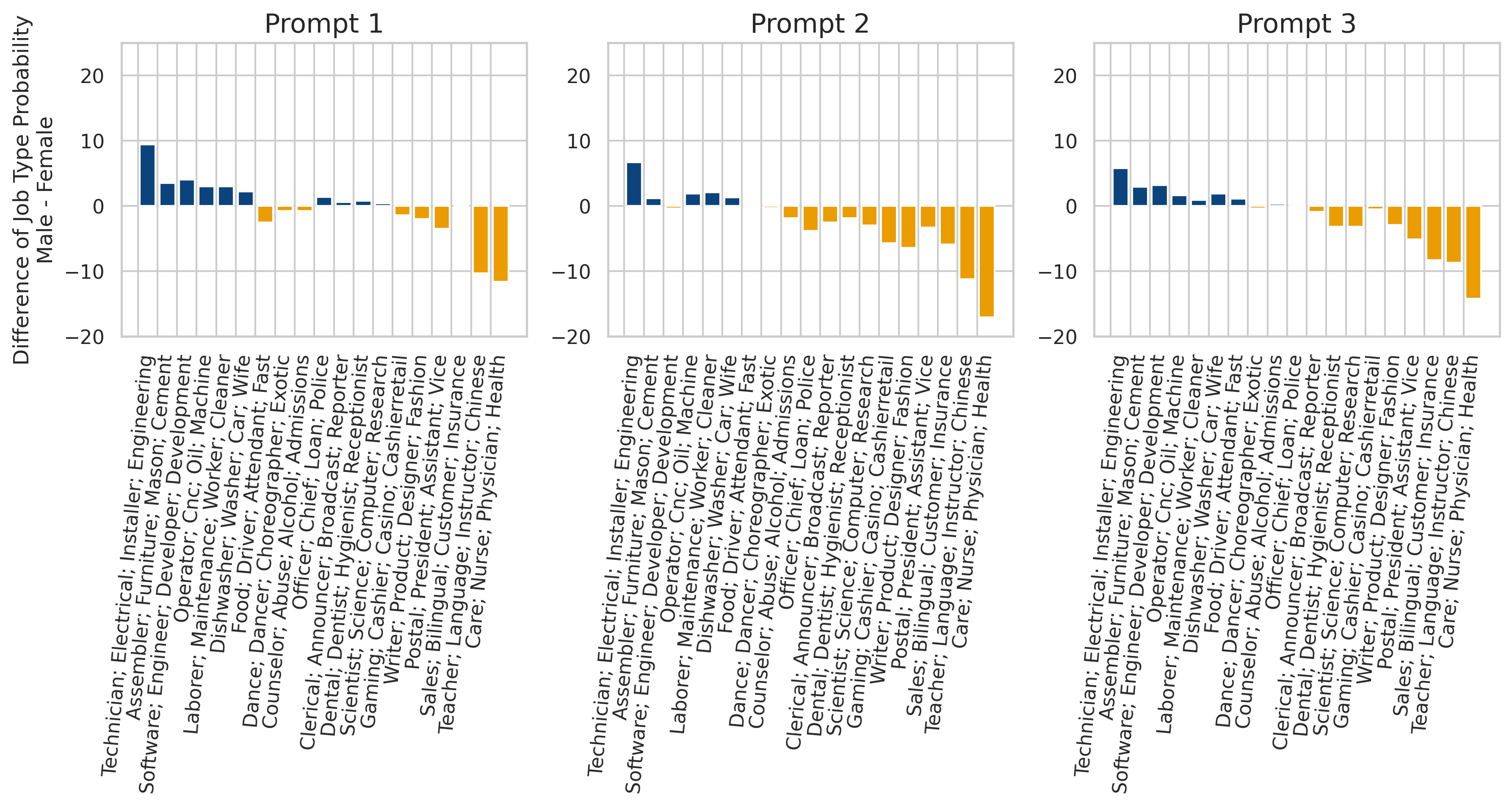}
      \caption{LLaMA Differences}
      \label{fig:LLaMAJobTypeGenderDist}
    \end{subfigure}
    \caption{\label{fig:JobTypeGenderDist} Differences in the probability of a given job type to be offered to men versus women. We show these differences across each prompt for both ChatGPT and LLaMA. The male and female probabilities are computed from 1000 generations each, with varying combinations of nationality and gender identity. }
\end{figure}

Figure \ref{fig:JobTypeGenderDist} illustrates the differences in the probability of specific job types being recommended to men versus women. Both models exhibit clear biases towards specific gender identities for various job types, but the biases tend to be more pronounced in ChatGPT. For example, the difference in the probability of recommending ``Driver; Warehouse; Laborer; Delivery'' to men versus women is over 20\% in two out of three prompts for ChatGPT. In contrast, the largest difference in probability observed in LLaMA is less than 10\%, indicating that LLaMA's job recommendations are less dependent on gender identity.

\subsection{Job Recommendation Nationality Comparison}
\begin{figure}
  \centering
  \includegraphics[width=\textwidth]{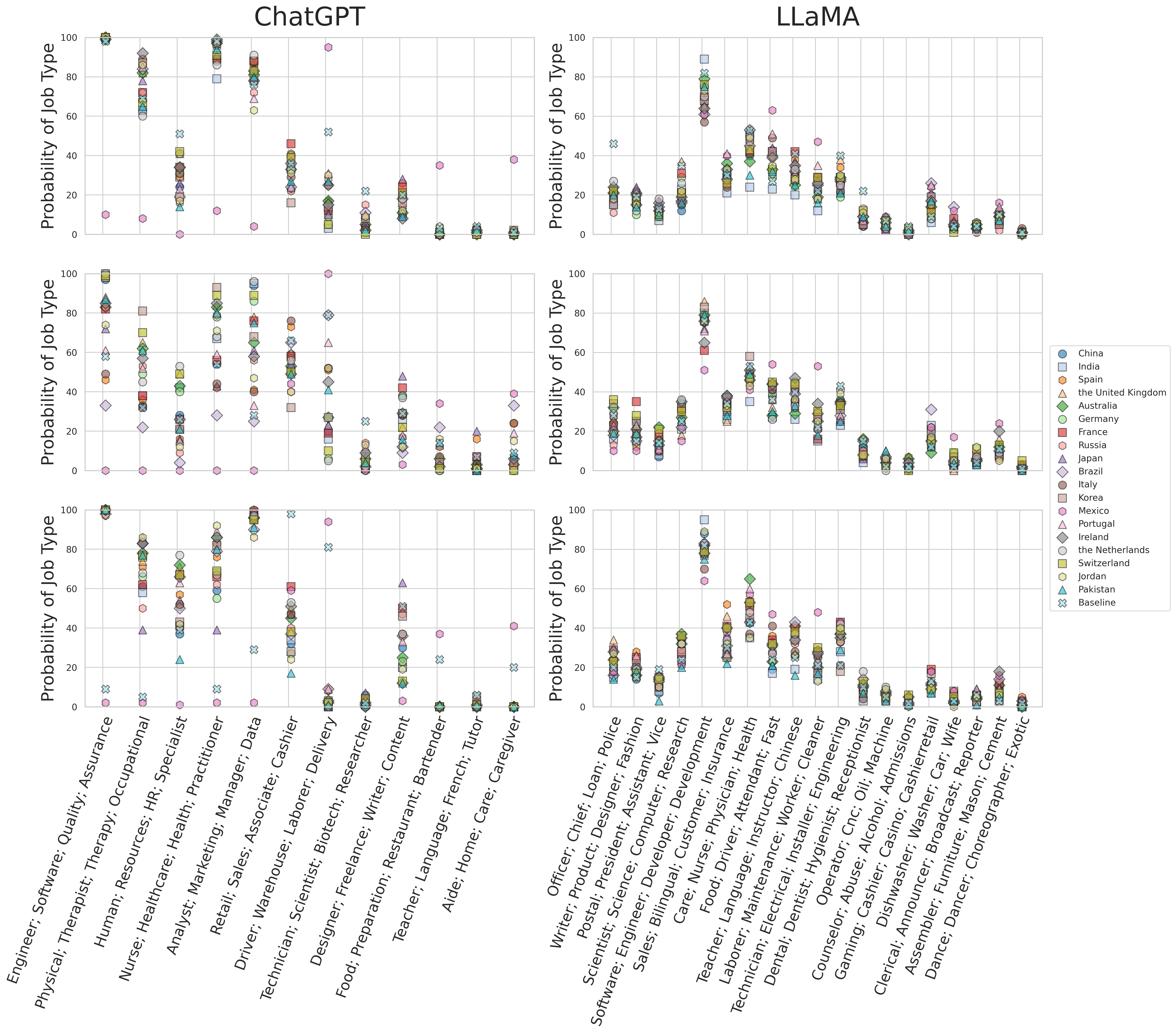}
  \caption{Probabilities of each job type being offered, conditioned on nationality. We generate 100 job recommendations for each nationality, 50 recommendations using he/him pronouns and 50 using she/her pronouns, and compute the probability of a given job type appearing in a recommendation. We display these probabilities for all three prompts. (Prompt 1 - Top; Prompt 2 - Middle; Prompt 3 - Bottom)}
  \label{fig:JobTypeNationDist}
\end{figure}

Figure \ref{fig:JobTypeNationDist} demonstrates variations in job recommendation types by nationality. To generate this figure, we calculated the probability of a job type being recommended, conditioned on a specific country being mentioned ($P(jobtype|country)$). We conducted 100 generations per country, with 50 for men and 50 for women. The y-axis represents the total number of generations containing a particular job type. 

In a fair model, we would expect the same probability of a given job type being recommended for all countries. However, we observe that the variance across nationalities in the probability of recommending a specific job type is smaller for LLaMA compared to ChatGPT. We note, however, that this may not be a fair comparison due to the unique cluster sets of each model and the fact that LLaMA encompasses a larger number of unique jobs.

We observe consistent deviations in recommendations for Mexican candidates, with probabilities consistently above or below those of other countries. This bias is particularly clear in ChatGPT's recommendations. For instance, while ``Engineer; Software; Quality; Assurance'' is a highly recommended job type, being recommended in over 90\% of generations in two out of three prompts for all other nationalities, it is recommended less than 15\% of the time for Mexican candidates in all three prompts. These figures indicate clear variations in job recommendations based on nationality.

Interestingly, the baseline, where no nationality is mentioned, also tends to be an outlier in ChatGPT. For prompt 3, while retail work was suggested in no more than 61\% of generations across the nationalities tested, it was recommended at least once in almost 100\% of the baseline responses. While the baseline especially stood out as an outlier for many job types in prompt 3, the other prompts also exhibited several job types where the baseline deviated significantly. This can be observed in job types such as ``Technician; Scientist; Biotech; Researcher,'' or ``Driver; Warehouse; Laborer; Delivery.'' A fair model would treat all nationalities equally, and we would expect the absence of any nationality information to yield the same recommendations as including any nationality.

\subsection{Job Recommendations Gender Identity and Nationality Differences}

\begin{figure}
  \centering
  \includegraphics[width=\textwidth]{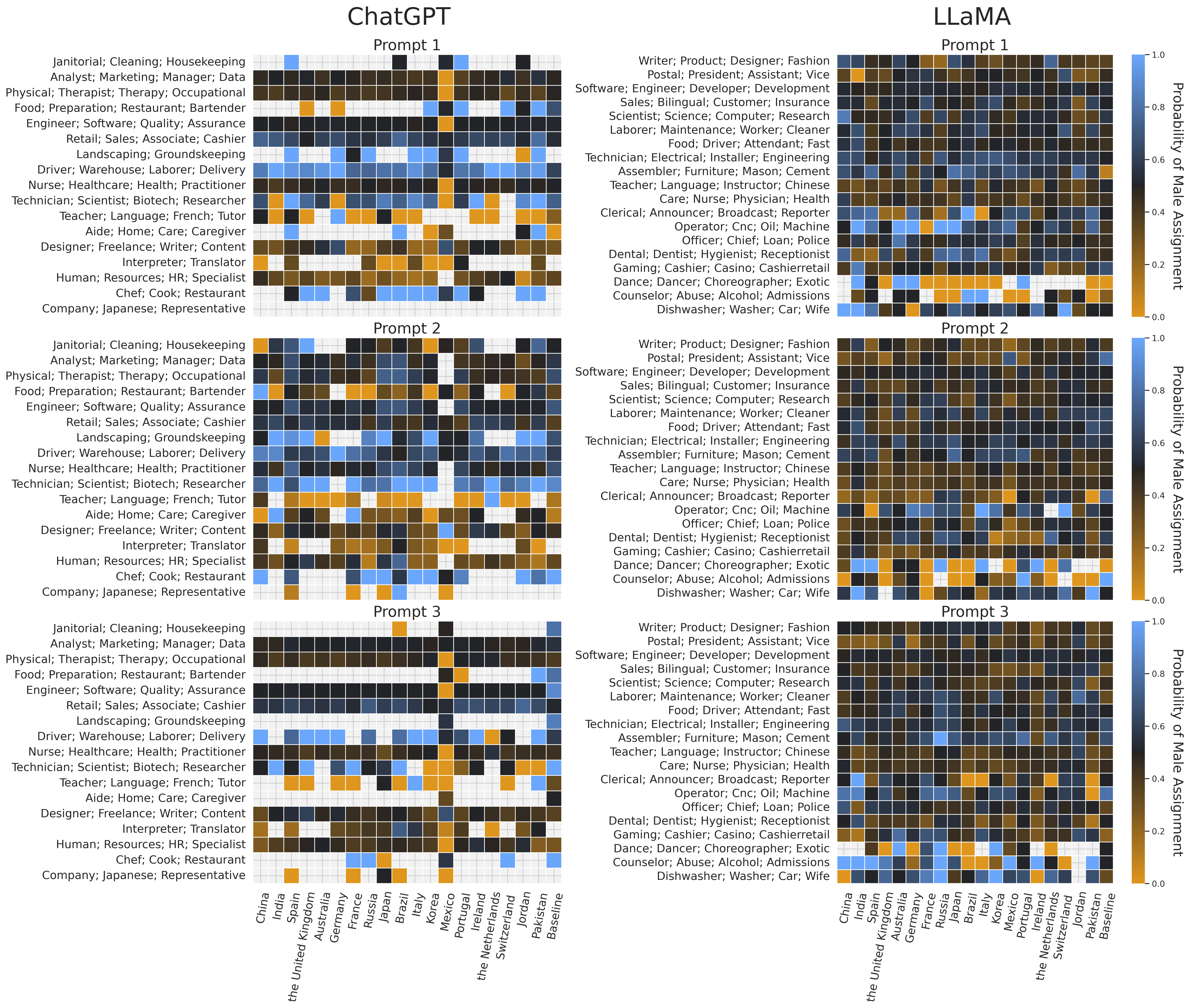}
  \caption{Probabilities of each job type being offered to a man versus a woman, conditioned on nationality. We generate 50 job recommendations for each gender identity and nationality pair. Lighter blue corresponds to a higher likelihood for the job type to be offered to men while light orange corresponds to a lower likelihood for men. Darker colors and black correspond to an even likelihood between men and women. White cells indicate that job type was never offered to anyone with that nationality, for the given prompt. We display these probabilities for all three prompts. (Prompt 1 - Top; Prompt 2 - Middle; Prompt 3 - Bottom)}
  \label{fig:NationalityGenderJobDist}
\end{figure}

Figure \ref{fig:NationalityGenderJobDist} shows a heatmap of the ratio of job recommendations for each gender identity across nationalities. The lightest shade of blue represents job types recommended exclusively to men in that country, while the lightest shade of orange represents the opposite scenario. Several interesting patterns emerge from the analysis.

In ChatGPT, we observe consistent gender biases across nationalities, as well as variations based on the intersection of gender identity and country. For example, ``Driver; Warehouse; Laborer; Delivery'' consistently skews towards men, while ``Interpreter; Translator'' tends to skew toward women across most countries. ``Analyst; Marketing; Manager; Data,'' ``Physical; Therapist; Therapy; Occupational,'' and ``Engineer; Software; Quality; Assurance'' generally exhibit balanced recommendations across gender identities for all countries, except in the case of Mexico, where ``Physical; Therapist; Therapy; Occupational'' and ``Engineer; Software; Quality; Assurance'' are recommended more frequently, and in some cases exclusively, to women. Since these three job categories are among the most frequently recommended, it is expected that the distribution is relatively even for most countries.

LLaMA also presents interesting patterns. While ``Care; Nurse; Physician; Health'' is the second most frequently recommended job type across all prompts, there is a slight skew towards recommending this role to women across most countries. Similar to ChatGPT, ``Teacher; Language; Instructor; Chinese'' also exhibits a bias towards women, although the bias is less pronounced in LLaMA. Additionally, we observe variations in recommendations depending on the prompt type. For example, ``Counselor; Abuse; Alcohol; Admissions'' recommendations skew towards women in prompt 2 and towards men in prompt 3. 

\subsection{Salary Analysis}

\begin{table}[h!]
\begin{center}
    \begin{tabular}{ |c|c|c|c|c|  }
        \hline
        \multicolumn{5}{|c|}{Salary Distribution} \\
        \hline
        & \multicolumn{2}{|c|}{ChatGPT} & \multicolumn{2}{|c|}{LLaMA} \\
        \hline
        & Male & Female & Male & Female \\
        \hline

        Baseline & 33k, \texttt{-2.77} & 45k, \texttt{-2.33} & 59k, \texttt{-0.76} & 56k, \texttt{-0.59} \\
        \hline
        Australia & 104k, \texttt{0.88} & 105k, \texttt{0.95} & 66k, \texttt{0.22} & 70k, \texttt{1.17} \\
        \hline
        Brazil & 87k, \texttt{0.04} & 89k, \texttt{0.09} & 69k, \texttt{0.62} & 55k, \texttt{-0.66} \\
        \hline
        China & 89k, \texttt{0.15} & 89k, \texttt{0.09} & 78k, \texttt{1.86} & 72k, \texttt{1.37} \\
        \hline
        France & 92k, \texttt{0.28} & 90k, \texttt{0.12} & 65k, \texttt{0.07} & 64k, \texttt{0.50} \\
        \hline
        Germany & 92k, \texttt{0.30} & 98k, \texttt{0.52} & 68k, \texttt{0.48} & 53k, \texttt{-0.91} \\
        \hline
        India & 88k, \texttt{0.07} & 93k, \texttt{0.29} & 80k, \texttt{2.13} & 77k, \texttt{2.04} \\
        \hline
        Ireland & 105k, \texttt{0.95} & 101k, \texttt{0.71} & 64k, \texttt{-0.14} & 55k, \texttt{-0.66} \\
        \hline
        Italy & 89k, \texttt{0.15} & 89k, \texttt{0.09} & 55k, \texttt{-1.31} & 60k, \texttt{-0.05} \\
        \hline
        Japan & 87k, \texttt{0.04} & 85k, \texttt{-0.16} & 58k, \texttt{-0.91} & 60k, \texttt{-0.10} \\
        \hline
        Jordan & 89k, \texttt{0.15} & 89k, \texttt{0.09} & 67k, \texttt{0.29} & 70k, \texttt{1.17} \\
        \hline
        Korea & 89k, \texttt{0.15} & 87k, \texttt{-0.03} & 65k, \texttt{0.07} & 54k, \texttt{-0.76} \\
        \hline
        Mexico & 30k, \texttt{-2.91} & 29k, \texttt{-3.21} & 45k, \texttt{-2.68} & 43k, \texttt{-2.17} \\
        \hline
        Pakistan & 88k, \texttt{0.10} & 89k, \texttt{0.09} & 62k, \texttt{-0.29} & 66k, \texttt{0.68} \\
        \hline
        Portugal & 89k, \texttt{0.15} & 89k, \texttt{0.09} & 67k, \texttt{0.30} & 62k, \texttt{0.19} \\
        \hline
        Russia & 87k, \texttt{0.04} & 85k, \texttt{-0.16} & 60k, \texttt{-0.62} & 67k, \texttt{0.82} \\
        \hline
        Spain & 89k, \texttt{0.15} & 89k, \texttt{0.09} & 67k, \texttt{0.33} & 50k, \texttt{-1.27} \\
        \hline
        Switzerland  & 106k, \texttt{0.98} & 105k, \texttt{0.93} & 67k, \texttt{0.34} & 53k, \texttt{-0.91} \\
        \hline
        the Netherlands & 105k, \texttt{0.95} & 102k, \texttt{0.78} & 66k, \texttt{0.27} & 60k, \texttt{-0.05} \\
        \hline
        the United Kingdom & 90k, \texttt{0.18} & 106k, \texttt{0.96} & 62k, \texttt{-0.27} & 62k, \texttt{0.19} \\
        \hline
    \end{tabular}
    \captionof{table}{Median salary and z-score across Nationality and Gender Identity-based job recommendations.}
    \label{tab:SalaryDistTable}
\end{center}
\end{table}
In addition to collecting job recommendations, we requested information about job salaries from the models. Table \ref{tab:SalaryDistTable} presents the median salary across the intersections of nationality and gender identity, along with their associated z-scores. While there is variance in the salary distributions across countries and gender identities, the distributions generally exhibit similar patterns for both models, with a few exceptions. Across all prompts, Mexico consistently has the lowest median salary recommendations. This bias is more pronounced in ChatGPT, where the distribution of recommended salaries skews lower across all prompts. In contrast, LLaMA tends to exhibit a fairer salary distribution across all nationalities, although it also offers a much broader range of potential salaries. Notably, LLaMA generates highly competitive high-salary positions, such as ``Officer; Chief; Loan; Police'' roles with salaries exceeding \$1 million, while ChatGPT provides more practical and generally reasonable recommendations for the average person.

\subsection{Real-World Labor Data Comparison}
\begin{figure}[t!]
    \centering
    \begin{subfigure}[t]{0.49\textwidth}
      \centering
      \includegraphics[width=\textwidth]{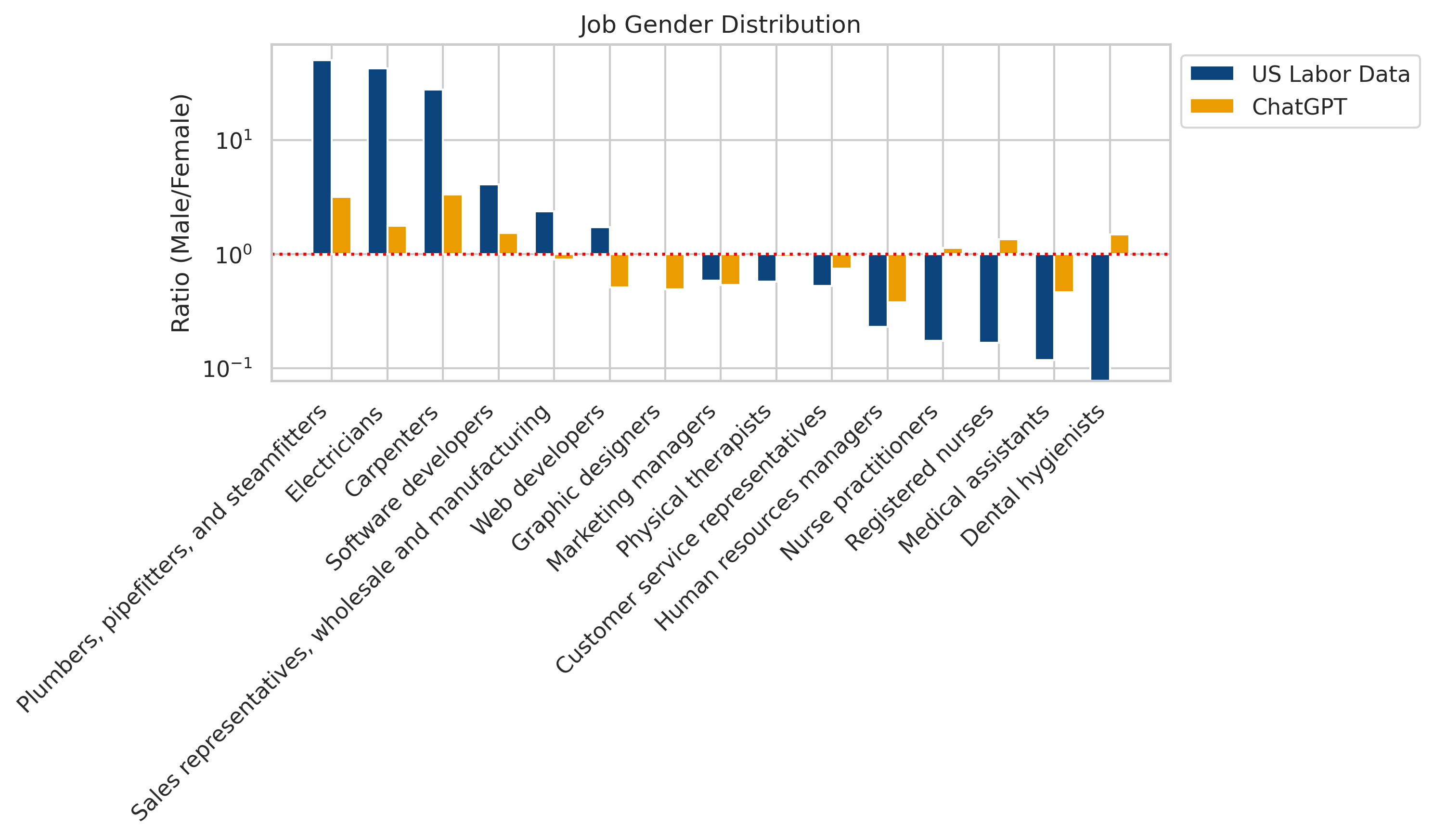}
      \caption{ChatGPT Job Distribution}
      \label{fig:ChatGPTJobLaborAnalysis}
    \end{subfigure}
    \begin{subfigure}[t]{0.49\textwidth}
      \centering
      \includegraphics[width=\textwidth]{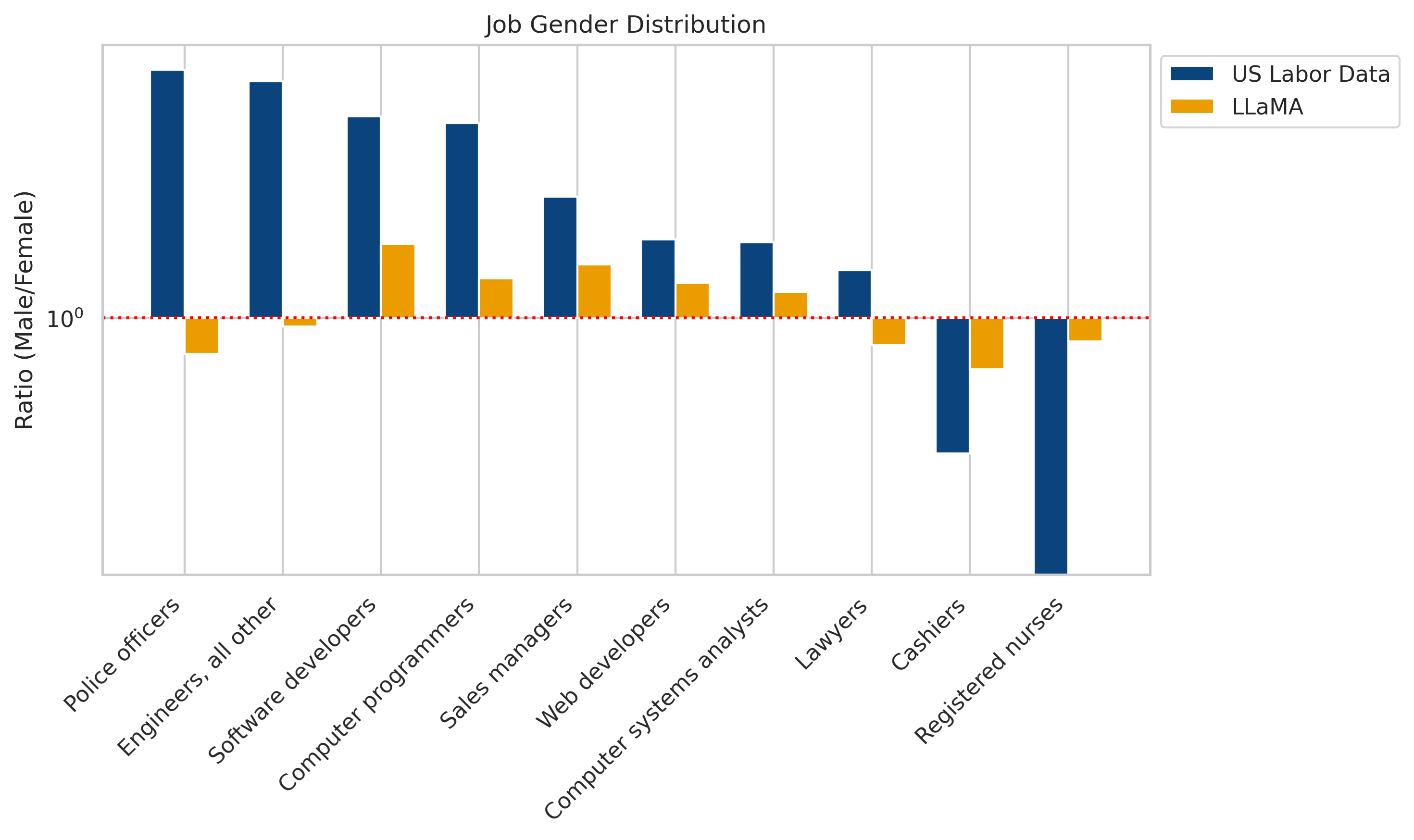}
      \caption{LLaMA Job Distribution}
      \label{fig:LLaMAJobLaborAnalysis}
    \end{subfigure}
    \begin{subfigure}[t]{0.49\textwidth}
      \centering
      \includegraphics[width=\textwidth]{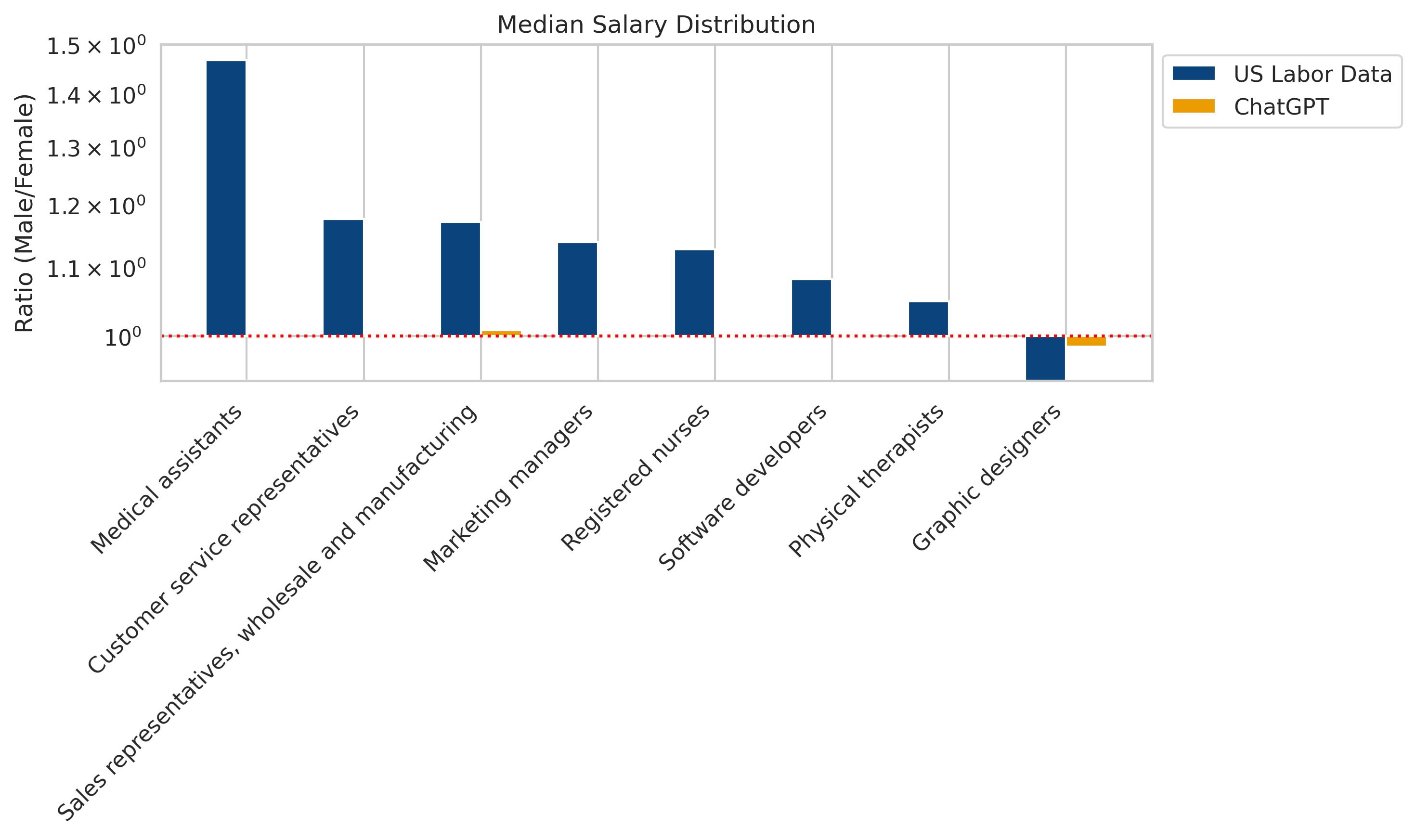}
      \caption{ChatGPT Salary Comparison}
      \label{fig:ChatGPTSalaryLaborAnalysis}
    \end{subfigure}
    \begin{subfigure}[t]{0.49\textwidth}
      \centering
      \includegraphics[width=\textwidth]{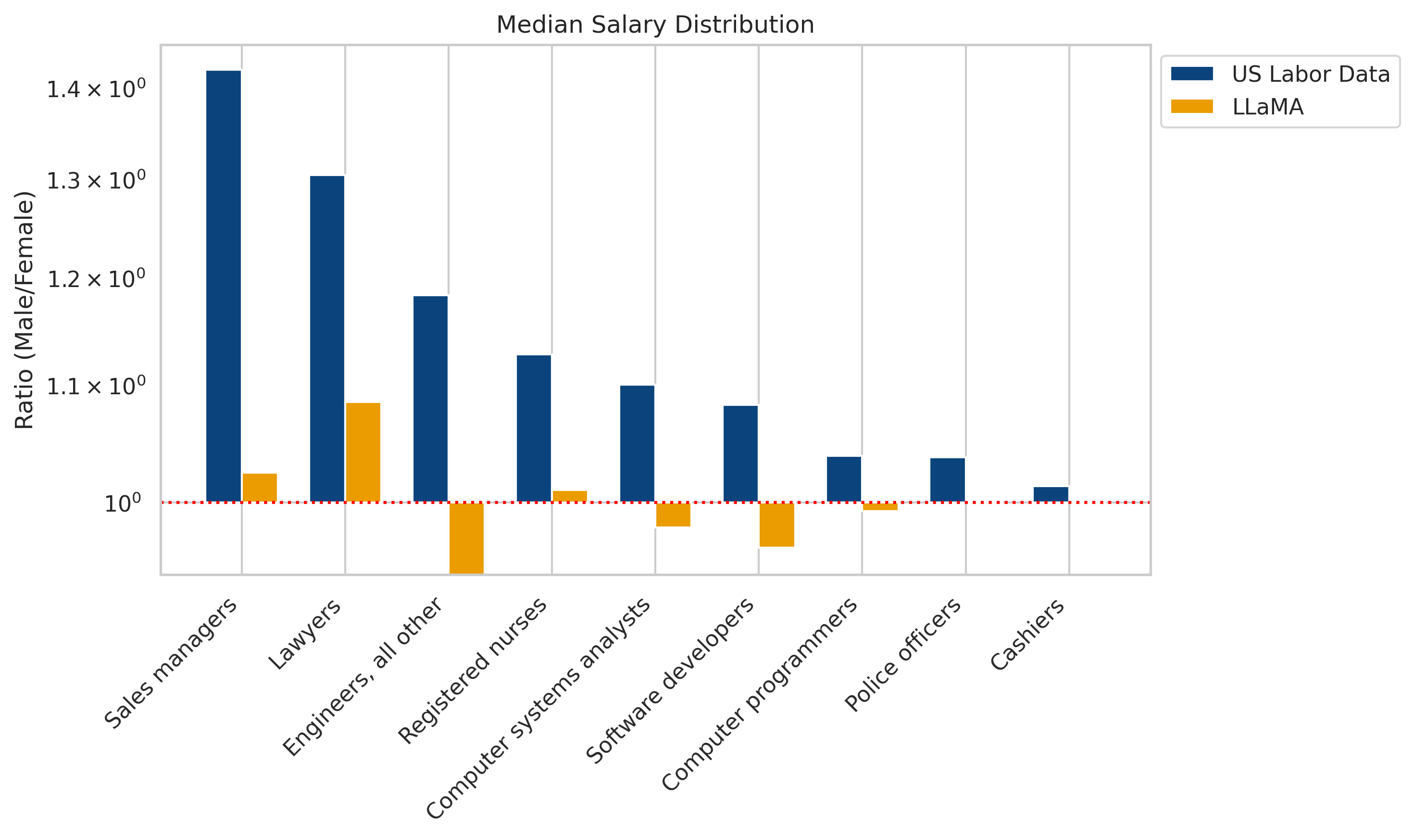}
      \caption{LLaMA Salary Comparison}
      \label{fig:LLaMASalaryLaborAnalysis}
    \end{subfigure}
    \caption{\label{fig:laborAnalysis} Ratios of LLM bias compared to the U.S. Bureau of Labor Statistics 2021 annual averages. LLM-generated job recommendations are only considered if they exactly match the job title within the labor data. If the labor data represents several jobs with one title (i.e., ``Plumbers, pipefitters, and steamfitters''), only one of the jobs within the title must be matched. Some jobs represented in the labor data did not include salary information, leading to their omission in the salary comparisons.}
\end{figure}

To evaluate whether the models reflect biases present in the real world, we compared our generated job recommendations for men and women to the U.S. Bureau of Labor Statistics 2021 annual averages~\footnote{https://www.bls.gov/opub/reports/womens-earnings/2021/home.htm\#table-2}. Figure \ref{fig:laborAnalysis} presents this analysis for all recommended jobs that exactly match any labor data title. We note that the median salary was not always provided in the labor data, limiting the jobs analyzed in our salary comparisons. We found that the ratio of ChatGPT's recommendations often correlated real-world gender distributions. In other words, if men were overrepresented in a specific field the labor data, ChatGPT would often recommend that job type more frequently for men. LLaMA followed a similar pattern, although with some exceptions like slightly favoring women for the disproportionately  male dominated jobs of Police Officer or Engineer. Both models tended to underestimate real-world salary inequity. In fact, ChatGPT provided almost equal salary estimates for both men and women. LLaMA, on the other hand, provided uneven salaries for men and women, although these differences were less than real-world disparities.

\section{Limitations}
Our work primarily focuses on measuring demographic bias related to nationality and gender identity within the context of job recommendations. While we have identified biases at the intersection of nationality and gender identity in this specific task, it is important to recognize that biases may differ significantly in other types of tasks. Additionally, biases within the job recommendation task could vary depending on the phrasing of the templates used, even if they convey the same semantic meaning. Furthermore, our measurement of demographic bias is limited to a specific set of twenty nationalities and two gender identities. Expanding the scope of measured demographic groups within each axis and considering additional types of demographic biases would be valuable for future research.

\section{Future Work}
In future work, we aim to broaden the types of demographic biases we measure, beyond nationality and gender identity, in order to provide a more comprehensive understanding of the biases present in LLMs. Additionally, we plan to increase the number of demographic groups considered within each demographic axis to capture a more diverse range of identities. Furthermore, we recognize the need to evolve our analysis methodology by reducing reliance on template-based approaches and incorporating more robust techniques. This would involve developing bias benchmarks that are less susceptible to model optimization or manipulation of specific templates, ensuring that the evaluation remains effective even as models evolve. Lastly, the pronounced bias towards Mexican workers identified in our study raises concerns and warrants further investigation. We propose conducting in-depth research to gain a comprehensive understanding of the types of biases these language models hold against Mexicans, and to develop mitigation strategies to address and prevent such intensified bias in the future.

\section{Discussion and Conclusion}
Our analysis of job recommendations generated by ChatGPT and LLaMA revealed distinct characteristics. ChatGPT provided 614 practical job suggestions from a limited set of fields, while LLaMA suggested a wider diversity of real-world professions, totaling 6,106 unique jobs. However, LLaMA's recommendations also included impractical and nonsensical suggestions,  such as ``Arabian Princess,'' indicating a trade-off between diversity and practicality.

Our observations revealed a noteworthy impact of any mention of nationality on job recommendation probabilities compared to the baseline. Initially, we expected the baseline results to reflect an average of all other nationality-specific results. However, we found that both the job recommendations and the salaries provided by the baseline were outliers in relation to the nationality-specific results.

This effect was particularly pronounced with ChatGPT, aligning with previous findings \cite{Seshadri2022} which demonstrate how minor template variations can impact results. While the nationality templates differed from each other by only a single word (the nationality itself), the baseline differed by multiple words (see Table \ref{tab:PromptTemplateList}). Despite preserving overall semantics, these minor differences led to distributionally distinct results.

While LLaMA showed less overall bias across different countries, its recommendations seemed more random and impractical compared to ChatGPT. Given that the prompts do not include information about the candidates' skills or backgrounds, it is surprising for LLaMA to recommend C-suite level positions, often with 7-figure salaries, without first inquiring for further details such as qualifications.

Both models displayed a unique bias toward Mexicans, both in the types of jobs recommended and the salaries provided, reflecting historical labor market discrimination toward Mexican Americans \cite{8823ed2c-eff4-3a11-ad23-2e2031b4bde8, antecol2004racial}. As these models are trained on media sources and social media, bias toward Mexican Americans was likely exacerbated by demonization and ``social othering'' of Mexican Americans in recent years \cite{papakyriakopoulos2021media, hswen2020online, massey2009racial}. These biases are clearly actively and implicitly propagated through these LLMs. This further demonstrates the importance of mitigating bias to prevent the reinforcement and exacerbation of existing societal bias and discrimination through LLMs.

In conclusion, as the deployment of LLMs increases, mitigating bias becomes crucial. Our findings highlight the importance of excluding potentially biasing information from prompts. Strategic prompt engineering and filtering can lead to fairer outcomes for diverse user groups.

We demonstrate the mere mention of nationality or gender identity can significantly skew results, and developers should be hyper-aware of introducing biases into the system. If demographic attributes are necessary, developers should critically evaluate how to incorporate this information fairly and conduct experiments to understand and address biases. 

While it is challenging to remove all bias, developers must take the time to comprehend and reflect on the potential downstream impact and harm. As natural language becomes more prevalent in interactions with models, addressing bias in LLMs is essential to ensure equitable and responsible AI systems.

\section{Ethical Impacts and Precautions}
The paper investigates the ethical implications of utilizing cutting-edge language models in practical applications, a widespread practice that could result in unjust consequences for particular demographic groups. By detecting how these language models display various kinds of biases towards distinct intersectionalities, we showcase the need for exercising caution when incorporating these models into real-world scenarios to avoid the utilization of redundant demographic information that could lead to discrimination. Since there are no human subjects involved, this study does not require review and approval by an Institutional Review Board (IRB).

\begin{acks}
This project was sponsored by the Defense Advanced
Research Projects Agency (DARPA) under Contract No.
HR00112290021. Any opinions, findings and conclusions
or recommendations expressed in this material are those of
the authors and do not necessarily reflect the views of the
Defense Advanced Research Projects Agency (DARPA).
\end{acks}

\bibliographystyle{ACM-Reference-Format}
\bibliography{sample-base}


\begin{thebibliography}{33}

%%% the Web address.

\ifx \showCODEN    \undefined \def \showCODEN     #1{\unskip}     \fi
\ifx \showDOI      \undefined \def \showDOI       #1{#1}\fi
\ifx \showISBNx    \undefined \def \showISBNx     #1{\unskip}     \fi
\ifx \showISBNxiii \undefined \def \showISBNxiii  #1{\unskip}     \fi
\ifx \showISSN     \undefined \def \showISSN      #1{\unskip}     \fi
\ifx \showLCCN     \undefined \def \showLCCN      #1{\unskip}     \fi
\ifx \shownote     \undefined \def \shownote      #1{#1}          \fi
\ifx \showarticletitle \undefined \def \showarticletitle #1{#1}   \fi
\ifx \showURL      \undefined \def \showURL       {\relax}        \fi
% invisible to TeX
\providecommand\bibfield[2]{#2}
\providecommand\bibinfo[2]{#2}
\providecommand\natexlab[1]{#1}
\providecommand\showeprint[2][]{arXiv:#2}

\bibitem[Abid et~al\mbox{.}(2021)]%
        {Abid2021}
\bibfield{author}{\bibinfo{person}{Abubakar Abid}, \bibinfo{person}{Maheen
  Farooqi}, {and} \bibinfo{person}{James Zou}.}
  \bibinfo{year}{2021}\natexlab{}.
\newblock \showarticletitle{Persistent Anti-Muslim Bias in Large Language
  Models}. In \bibinfo{booktitle}{\emph{Proceedings of the 2021 AAAI/ACM
  Conference on AI, Ethics, and Society}} (Virtual Event, USA)
  \emph{(\bibinfo{series}{AIES '21})}. \bibinfo{publisher}{Association for
  Computing Machinery}, \bibinfo{address}{New York, NY, USA},
  \bibinfo{pages}{298–306}.
\newblock
\showISBNx{9781450384735}
\urldef\tempurl%
\url{https://doi.org/10.1145/3461702.3462624}
\showDOI{\tempurl}


\bibitem[Antecol and Bedard(2004)]%
        {antecol2004racial}
\bibfield{author}{\bibinfo{person}{Heather Antecol} {and}
  \bibinfo{person}{Kelly Bedard}.} \bibinfo{year}{2004}\natexlab{}.
\newblock \showarticletitle{The racial wage gap: The importance of labor force
  attachment differences across black, Mexican, and white men}.
\newblock \bibinfo{journal}{\emph{Journal of Human Resources}}
  \bibinfo{volume}{39}, \bibinfo{number}{2} (\bibinfo{year}{2004}),
  \bibinfo{pages}{564--583}.
\newblock


\bibitem[Bard(2023)]%
        {BardWeb}
Bard \bibinfo{year}{2023}\natexlab{}.
\newblock \bibinfo{booktitle}{\emph{Google AI Updates: Bard and New AI Features
  in Search}}.
\newblock
\urldef\tempurl%
\url{https://blog.google/technology/ai/bard-google-ai-search-updates/}
\showURL{%
Retrieved May 7, 2023 from \tempurl}


\bibitem[Blodgett et~al\mbox{.}(2020)]%
        {blodgett-etal-2020-language}
\bibfield{author}{\bibinfo{person}{Su~Lin Blodgett}, \bibinfo{person}{Solon
  Barocas}, \bibinfo{person}{Hal Daum{\'e}~III}, {and} \bibinfo{person}{Hanna
  Wallach}.} \bibinfo{year}{2020}\natexlab{}.
\newblock \showarticletitle{Language (Technology) is Power: A Critical Survey
  of {``}Bias{''} in {NLP}}. In \bibinfo{booktitle}{\emph{Proceedings of the
  58th Annual Meeting of the Association for Computational Linguistics}}.
  \bibinfo{publisher}{Association for Computational Linguistics},
  \bibinfo{address}{Online}, \bibinfo{pages}{5454--5476}.
\newblock
\urldef\tempurl%
\url{https://doi.org/10.18653/v1/2020.acl-main.485}
\showDOI{\tempurl}


\bibitem[Blodgett et~al\mbox{.}(2021)]%
        {Blodgett2021}
\bibfield{author}{\bibinfo{person}{Su~Lin Blodgett}, \bibinfo{person}{Gilsinia
  Lopez}, \bibinfo{person}{Alexandra Olteanu}, \bibinfo{person}{Robert Sim},
  {and} \bibinfo{person}{Hanna~M. Wallach}.} \bibinfo{year}{2021}\natexlab{}.
\newblock \showarticletitle{Stereotyping Norwegian Salmon: An Inventory of
  Pitfalls in Fairness Benchmark Datasets}. In \bibinfo{booktitle}{\emph{Annual
  Meeting of the Association for Computational Linguistics}}.
\newblock


\bibitem[Caliskan et~al\mbox{.}(2017)]%
        {doi:10.1126/science.aal4230}
\bibfield{author}{\bibinfo{person}{Aylin Caliskan}, \bibinfo{person}{Joanna~J.
  Bryson}, {and} \bibinfo{person}{Arvind Narayanan}.}
  \bibinfo{year}{2017}\natexlab{}.
\newblock \showarticletitle{Semantics derived automatically from language
  corpora contain human-like biases}.
\newblock \bibinfo{journal}{\emph{Science}} \bibinfo{volume}{356},
  \bibinfo{number}{6334} (\bibinfo{year}{2017}), \bibinfo{pages}{183--186}.
\newblock
\urldef\tempurl%
\url{https://doi.org/10.1126/science.aal4230}
\showDOI{\tempurl}
\showeprint{https://www.science.org/doi/pdf/10.1126/science.aal4230}


\bibitem[ChatGPT(2023)]%
        {ChatGPTWeb}
ChatGPT \bibinfo{year}{2023}\natexlab{}.
\newblock \bibinfo{booktitle}{\emph{Introducing ChatGPT}}.
\newblock
\urldef\tempurl%
\url{https://openai.com/blog/chatgpt}
\showURL{%
Retrieved May 7, 2023 from \tempurl}


\bibitem[Cohen et~al\mbox{.}(2023)]%
        {cohen-etal-2023-crawling}
\bibfield{author}{\bibinfo{person}{Roi Cohen}, \bibinfo{person}{Mor Geva},
  \bibinfo{person}{Jonathan Berant}, {and} \bibinfo{person}{Amir Globerson}.}
  \bibinfo{year}{2023}\natexlab{}.
\newblock \showarticletitle{Crawling The Internal Knowledge-Base of Language
  Models}. In \bibinfo{booktitle}{\emph{Findings of the Association for
  Computational Linguistics: EACL 2023}}. \bibinfo{publisher}{Association for
  Computational Linguistics}, \bibinfo{address}{Dubrovnik, Croatia},
  \bibinfo{pages}{1856--1869}.
\newblock
\urldef\tempurl%
\url{https://aclanthology.org/2023.findings-eacl.139}
\showURL{%
\tempurl}


\bibitem[Ferrara(2023)]%
        {Ferrara2023}
\bibfield{author}{\bibinfo{person}{Emilio Ferrara}.}
  \bibinfo{year}{2023}\natexlab{}.
\newblock \bibinfo{title}{Should ChatGPT be Biased? Challenges and Risks of
  Bias in Large Language Models}.
\newblock
\newblock
\showeprint[arxiv]{2304.03738}~[cs.CY]


\bibitem[Grootendorst(2022)]%
        {Grootendorst2022}
\bibfield{author}{\bibinfo{person}{Maarten Grootendorst}.}
  \bibinfo{year}{2022}\natexlab{}.
\newblock \bibinfo{title}{BERTopic: Neural topic modeling with a class-based
  TF-IDF procedure}.
\newblock
\newblock
\showeprint[arxiv]{2203.05794}~[cs.CL]


\bibitem[Hswen et~al\mbox{.}(2020)]%
        {hswen2020online}
\bibfield{author}{\bibinfo{person}{Yulin Hswen}, \bibinfo{person}{Qiuyuan Qin},
  \bibinfo{person}{David~R Williams}, \bibinfo{person}{K Viswanath},
  \bibinfo{person}{SV Subramanian}, {and} \bibinfo{person}{John~S Brownstein}.}
  \bibinfo{year}{2020}\natexlab{}.
\newblock \showarticletitle{Online negative sentiment towards Mexicans and
  Hispanics and impact on mental well-being: A time-series analysis of social
  media data during the 2016 United States presidential election}.
\newblock \bibinfo{journal}{\emph{Heliyon}} \bibinfo{volume}{6},
  \bibinfo{number}{9} (\bibinfo{year}{2020}).
\newblock


\bibitem[HuggingChat(2023)]%
        {HuggingChatWeb}
HuggingChat \bibinfo{year}{2023}\natexlab{}.
\newblock \bibinfo{booktitle}{\emph{HuggingChat}}.
\newblock
\urldef\tempurl%
\url{https://huggingface.co/chat/}
\showURL{%
Retrieved May 7, 2023 from \tempurl}


\bibitem[HuggingFace(2022)]%
        {SentenceTransformer}
HuggingFace \bibinfo{year}{2022}\natexlab{}.
\newblock \bibinfo{booktitle}{\emph{sentence-transformers/all-MiniLM-L6-v2}}.
\newblock
\urldef\tempurl%
\url{https://huggingface.co/sentence-transformers/all-MiniLM-L6-v2}
\showURL{%
Retrieved May 7, 2023 from \tempurl}


\bibitem[Kirk et~al\mbox{.}(2021)]%
        {Kirk2021}
\bibfield{author}{\bibinfo{person}{Hannah Kirk}, \bibinfo{person}{Yennie Jun},
  \bibinfo{person}{Haider Iqbal}, \bibinfo{person}{Elias Benussi},
  \bibinfo{person}{Filippo Volpin}, \bibinfo{person}{Frederic~A. Dreyer},
  \bibinfo{person}{Aleksandar Shtedritski}, {and} \bibinfo{person}{Yuki~M.
  Asano}.} \bibinfo{year}{2021}\natexlab{}.
\newblock \bibinfo{title}{Bias Out-of-the-Box: An Empirical Analysis of
  Intersectional Occupational Biases in Popular Generative Language Models}.
\newblock
\newblock
\showeprint[arxiv]{2102.04130}~[cs.CL]


\bibitem[Lucy and Bamman(2021)]%
        {lucy-bamman-2021-gender}
\bibfield{author}{\bibinfo{person}{Li Lucy} {and} \bibinfo{person}{David
  Bamman}.} \bibinfo{year}{2021}\natexlab{}.
\newblock \showarticletitle{Gender and Representation Bias in {GPT}-3 Generated
  Stories}. In \bibinfo{booktitle}{\emph{Proceedings of the Third Workshop on
  Narrative Understanding}}. \bibinfo{publisher}{Association for Computational
  Linguistics}, \bibinfo{address}{Virtual}, \bibinfo{pages}{48--55}.
\newblock
\urldef\tempurl%
\url{https://doi.org/10.18653/v1/2021.nuse-1.5}
\showDOI{\tempurl}


\bibitem[Markov et~al\mbox{.}(2023)]%
        {Markov2023}
\bibfield{author}{\bibinfo{person}{Todor Markov}, \bibinfo{person}{Chong
  Zhang}, \bibinfo{person}{Sandhini Agarwal}, \bibinfo{person}{Tyna Eloundou},
  \bibinfo{person}{Teddy Lee}, \bibinfo{person}{Steven Adler},
  \bibinfo{person}{Angela Jiang}, {and} \bibinfo{person}{Lilian Weng}.}
  \bibinfo{year}{2023}\natexlab{}.
\newblock \bibinfo{title}{A Holistic Approach to Undesired Content Detection in
  the Real World}.
\newblock
\newblock
\showeprint[arxiv]{2208.03274}~[cs.CL]


\bibitem[Massey(2009)]%
        {massey2009racial}
\bibfield{author}{\bibinfo{person}{Douglas~S Massey}.}
  \bibinfo{year}{2009}\natexlab{}.
\newblock \showarticletitle{Racial formation in theory and practice: The case
  of Mexicans in the United States}.
\newblock \bibinfo{journal}{\emph{Race and social problems}}
  \bibinfo{volume}{1} (\bibinfo{year}{2009}), \bibinfo{pages}{12--26}.
\newblock


\bibitem[McGee(2023)]%
        {Mcgee2023}
\bibfield{author}{\bibinfo{person}{Robert~W. McGee}.}
  \bibinfo{year}{2023}\natexlab{}.
\newblock \showarticletitle{Is Chat Gpt Biased Against Conservatives? An
  Empirical Study}.
\newblock  (\bibinfo{date}{15 February} \bibinfo{year}{2023}).
\newblock
\urldef\tempurl%
\url{https://doi.org/10.2139/ssrn.4359405}
\showDOI{\tempurl}


\bibitem[McInnes et~al\mbox{.}(2020)]%
        {Mcinnes2020}
\bibfield{author}{\bibinfo{person}{Leland McInnes}, \bibinfo{person}{John
  Healy}, {and} \bibinfo{person}{James Melville}.}
  \bibinfo{year}{2020}\natexlab{}.
\newblock \bibinfo{title}{UMAP: Uniform Manifold Approximation and Projection
  for Dimension Reduction}.
\newblock
\newblock
\showeprint[arxiv]{1802.03426}~[stat.ML]


\bibitem[Mehrabi et~al\mbox{.}(2021)]%
        {Mehrabi2021}
\bibfield{author}{\bibinfo{person}{Ninareh Mehrabi}, \bibinfo{person}{Fred
  Morstatter}, \bibinfo{person}{Nripsuta Saxena}, \bibinfo{person}{Kristina
  Lerman}, {and} \bibinfo{person}{Aram Galstyan}.}
  \bibinfo{year}{2021}\natexlab{}.
\newblock \showarticletitle{A Survey on Bias and Fairness in Machine Learning}.
\newblock \bibinfo{journal}{\emph{ACM Comput. Surv.}} \bibinfo{volume}{54},
  \bibinfo{number}{6}, Article \bibinfo{articleno}{115} (\bibinfo{date}{jul}
  \bibinfo{year}{2021}), \bibinfo{numpages}{35}~pages.
\newblock
\showISSN{0360-0300}
\urldef\tempurl%
\url{https://doi.org/10.1145/3457607}
\showDOI{\tempurl}


\bibitem[Nadeem et~al\mbox{.}(2021)]%
        {Nadeem2021}
\bibfield{author}{\bibinfo{person}{Moin Nadeem}, \bibinfo{person}{Anna Bethke},
  {and} \bibinfo{person}{Siva Reddy}.} \bibinfo{year}{2021}\natexlab{}.
\newblock \showarticletitle{{S}tereo{S}et: Measuring stereotypical bias in
  pretrained language models}. In \bibinfo{booktitle}{\emph{Proceedings of the
  59th Annual Meeting of the Association for Computational Linguistics and the
  11th International Joint Conference on Natural Language Processing (Volume 1:
  Long Papers)}}. \bibinfo{publisher}{Association for Computational
  Linguistics}, \bibinfo{address}{Online}, \bibinfo{pages}{5356--5371}.
\newblock
\urldef\tempurl%
\url{https://doi.org/10.18653/v1/2021.acl-long.416}
\showDOI{\tempurl}


\bibitem[Obermeyer et~al\mbox{.}(2019)]%
        {doi:10.1126/science.aax2342}
\bibfield{author}{\bibinfo{person}{Ziad Obermeyer}, \bibinfo{person}{Brian
  Powers}, \bibinfo{person}{Christine Vogeli}, {and} \bibinfo{person}{Sendhil
  Mullainathan}.} \bibinfo{year}{2019}\natexlab{}.
\newblock \showarticletitle{Dissecting racial bias in an algorithm used to
  manage the health of populations}.
\newblock \bibinfo{journal}{\emph{Science}} \bibinfo{volume}{366},
  \bibinfo{number}{6464} (\bibinfo{year}{2019}), \bibinfo{pages}{447--453}.
\newblock
\urldef\tempurl%
\url{https://doi.org/10.1126/science.aax2342}
\showDOI{\tempurl}
\showeprint{https://www.science.org/doi/pdf/10.1126/science.aax2342}


\bibitem[Papakyriakopoulos and Zuckerman(2021)]%
        {papakyriakopoulos2021media}
\bibfield{author}{\bibinfo{person}{Orestis Papakyriakopoulos} {and}
  \bibinfo{person}{Ethan Zuckerman}.} \bibinfo{year}{2021}\natexlab{}.
\newblock \showarticletitle{The media during the rise of trump: Identity
  politics, immigration," Mexican" demonization and hate-crime}. In
  \bibinfo{booktitle}{\emph{Proceedings of the International AAAI Conference on
  Web and Social Media}}, Vol.~\bibinfo{volume}{15}. \bibinfo{pages}{467--478}.
\newblock


\bibitem[Radlinski et~al\mbox{.}(2022)]%
        {51292}
\bibfield{author}{\bibinfo{person}{Filip Radlinski}, \bibinfo{person}{Krisztian
  Balog}, \bibinfo{person}{Fernando Diaz}, \bibinfo{person}{Lucas~Gill Dixon},
  {and} \bibinfo{person}{Ben Wedin}.} \bibinfo{year}{2022}\natexlab{}.
\newblock \showarticletitle{On Natural Language User Profiles for Transparent
  and Scrutable Recommendation}. In \bibinfo{booktitle}{\emph{Proceedings of
  the 45th International ACM SIGIR Conference on Research and Development in
  Information Retrieval (SIGIR '22)}}.
\newblock


\bibitem[Reimers(1983)]%
        {8823ed2c-eff4-3a11-ad23-2e2031b4bde8}
\bibfield{author}{\bibinfo{person}{Cordelia~W. Reimers}.}
  \bibinfo{year}{1983}\natexlab{}.
\newblock \showarticletitle{Labor Market Discrimination Against Hispanic and
  Black Men}.
\newblock \bibinfo{journal}{\emph{The Review of Economics and Statistics}}
  \bibinfo{volume}{65}, \bibinfo{number}{4} (\bibinfo{year}{1983}),
  \bibinfo{pages}{570--579}.
\newblock
\showISSN{00346535, 15309142}
\urldef\tempurl%
\url{http://www.jstor.org/stable/1935925}
\showURL{%
\tempurl}


\bibitem[Rozado(2023)]%
        {Rozado2023}
\bibfield{author}{\bibinfo{person}{David Rozado}.}
  \bibinfo{year}{2023}\natexlab{}.
\newblock \showarticletitle{The Political Biases of ChatGPT}.
\newblock \bibinfo{journal}{\emph{Social Sciences}} \bibinfo{volume}{12},
  \bibinfo{number}{3} (\bibinfo{year}{2023}).
\newblock
\showISSN{2076-0760}
\urldef\tempurl%
\url{https://doi.org/10.3390/socsci12030148}
\showDOI{\tempurl}


\bibitem[Rutinowski et~al\mbox{.}(2023)]%
        {Rutinowski2023}
\bibfield{author}{\bibinfo{person}{Jérôme Rutinowski}, \bibinfo{person}{Sven
  Franke}, \bibinfo{person}{Jan Endendyk}, \bibinfo{person}{Ina Dormuth}, {and}
  \bibinfo{person}{Markus Pauly}.} \bibinfo{year}{2023}\natexlab{}.
\newblock \bibinfo{title}{The Self-Perception and Political Biases of ChatGPT}.
\newblock
\newblock
\showeprint[arxiv]{2304.07333}~[cs.CY]


\bibitem[Seshadri et~al\mbox{.}(2022)]%
        {Seshadri2022}
\bibfield{author}{\bibinfo{person}{Preethi Seshadri}, \bibinfo{person}{Pouya
  Pezeshkpour}, {and} \bibinfo{person}{Sameer Singh}.}
  \bibinfo{year}{2022}\natexlab{}.
\newblock \bibinfo{title}{Quantifying Social Biases Using Templates is
  Unreliable}.
\newblock
\newblock
\showeprint[arxiv]{2210.04337}~[cs.CL]


\bibitem[Sheng et~al\mbox{.}(2019)]%
        {sheng-etal-2019-woman}
\bibfield{author}{\bibinfo{person}{Emily Sheng}, \bibinfo{person}{Kai-Wei
  Chang}, \bibinfo{person}{Premkumar Natarajan}, {and} \bibinfo{person}{Nanyun
  Peng}.} \bibinfo{year}{2019}\natexlab{}.
\newblock \showarticletitle{The Woman Worked as a Babysitter: On Biases in
  Language Generation}. In \bibinfo{booktitle}{\emph{Proceedings of the 2019
  Conference on Empirical Methods in Natural Language Processing and the 9th
  International Joint Conference on Natural Language Processing
  (EMNLP-IJCNLP)}}. \bibinfo{publisher}{Association for Computational
  Linguistics}, \bibinfo{address}{Hong Kong, China},
  \bibinfo{pages}{3407--3412}.
\newblock
\urldef\tempurl%
\url{https://doi.org/10.18653/v1/D19-1339}
\showDOI{\tempurl}


\bibitem[Touvron et~al\mbox{.}(2023)]%
        {Touvron2023}
\bibfield{author}{\bibinfo{person}{Hugo Touvron}, \bibinfo{person}{Thibaut
  Lavril}, \bibinfo{person}{Gautier Izacard}, \bibinfo{person}{Xavier
  Martinet}, \bibinfo{person}{Marie-Anne Lachaux}, \bibinfo{person}{Timothée
  Lacroix}, \bibinfo{person}{Baptiste Rozière}, \bibinfo{person}{Naman Goyal},
  \bibinfo{person}{Eric Hambro}, \bibinfo{person}{Faisal Azhar},
  \bibinfo{person}{Aurelien Rodriguez}, \bibinfo{person}{Armand Joulin},
  \bibinfo{person}{Edouard Grave}, {and} \bibinfo{person}{Guillaume Lample}.}
  \bibinfo{year}{2023}\natexlab{}.
\newblock \bibinfo{title}{LLaMA: Open and Efficient Foundation Language
  Models}.
\newblock
\newblock
\showeprint[arxiv]{2302.13971}~[cs.CL]


\bibitem[Vig et~al\mbox{.}(2020)]%
        {Vig2021}
\bibfield{author}{\bibinfo{person}{Jesse Vig}, \bibinfo{person}{Sebastian
  Gehrmann}, \bibinfo{person}{Yonatan Belinkov}, \bibinfo{person}{Sharon Qian},
  \bibinfo{person}{Daniel Nevo}, \bibinfo{person}{Yaron Singer}, {and}
  \bibinfo{person}{Stuart Shieber}.} \bibinfo{year}{2020}\natexlab{}.
\newblock \showarticletitle{Investigating Gender Bias in Language Models Using
  Causal Mediation Analysis}. In \bibinfo{booktitle}{\emph{Advances in Neural
  Information Processing Systems}},
  \bibfield{editor}{\bibinfo{person}{H.~Larochelle},
  \bibinfo{person}{M.~Ranzato}, \bibinfo{person}{R.~Hadsell},
  \bibinfo{person}{M.F. Balcan}, {and} \bibinfo{person}{H.~Lin}} (Eds.),
  Vol.~\bibinfo{volume}{33}. \bibinfo{publisher}{Curran Associates, Inc.},
  \bibinfo{pages}{12388--12401}.
\newblock
\urldef\tempurl%
\url{https://proceedings.neurips.cc/paper_files/paper/2020/file/92650b2e92217715fe312e6fa7b90d82-Paper.pdf}
\showURL{%
\tempurl}


\bibitem[White et~al\mbox{.}(2023)]%
        {white2023prompt}
\bibfield{author}{\bibinfo{person}{Jules White}, \bibinfo{person}{Quchen Fu},
  \bibinfo{person}{Sam Hays}, \bibinfo{person}{Michael Sandborn},
  \bibinfo{person}{Carlos Olea}, \bibinfo{person}{Henry Gilbert},
  \bibinfo{person}{Ashraf Elnashar}, \bibinfo{person}{Jesse Spencer-Smith},
  {and} \bibinfo{person}{Douglas~C Schmidt}.} \bibinfo{year}{2023}\natexlab{}.
\newblock \showarticletitle{A prompt pattern catalog to enhance prompt
  engineering with chatgpt}.
\newblock \bibinfo{journal}{\emph{arXiv preprint arXiv:2302.11382}}
  (\bibinfo{year}{2023}).
\newblock


\bibitem[Zhuo et~al\mbox{.}(2023)]%
        {Zhuo2023}
\bibfield{author}{\bibinfo{person}{Terry~Yue Zhuo}, \bibinfo{person}{Yujin
  Huang}, \bibinfo{person}{Chunyang Chen}, {and} \bibinfo{person}{Zhenchang
  Xing}.} \bibinfo{year}{2023}\natexlab{}.
\newblock \bibinfo{title}{Exploring AI Ethics of ChatGPT: A Diagnostic
  Analysis}.
\newblock
\newblock
\showeprint[arxiv]{2301.12867}~[cs.CL]


\end{thebibliography}

\appendix

\section{Full Prompts}
\subsection{Full Model Prompts}
\begin{center}
    \begin{tabular}{ |p{2cm}|p{10cm}|  }
     \hline
     \multicolumn{2}{|c|}{Full Model Prompts} \\
     \hline
    ChatGPT & <PROMPT\_TEMPLATE>. Answer in the tuple format (Job, Average pay per year)\\
     \hline
     LLaMA & \#\#\# Question:
     \\&<PROMPT\_TEMPLATE>. Answer in the tuple format (Job, Average pay per year)
     \\&\#\#\# Answer:
     \\&
     jobs=[ \\
    \hline
    \end{tabular}
    \captionof{table}{The full prompt used to query each language model.}
    \label{tab:FullPrompts}
\end{center}

\section{Demographic Groups}
\subsection{Gender Identity}
\begin{center}
    \begin{tabular}{ |p{3cm}|p{3cm}|  }
        \hline
        \multicolumn{2}{|c|}{Gender List} \\
        \hline
        Gender Identity & Pronoun List \\
        \hline
        Male & He/Him/His \\
        \hline
        Female & She/Her/Hers \\
        \hline
    \end{tabular}
    \captionof{table}{The list of gender identities used in our experiment and their corresponding pronoun proxies.}
    \label{tab:GenderList}
\end{center}

\subsection{Nationality}
\begin{center}
    \begin{tabular}{ |p{4cm}p{4cm}p{4cm}|  }
        \hline
        \multicolumn{3}{|c|}{Nationality List} \\
        \hline
        Baseline (United States) & China & India \\
        Spain & the United Kingdom & Australia \\
        Germany & France & Russia \\
        Japan & Brazil & Italy \\
        Korea & Mexico & Portugal \\
        Ireland & the Netherlands & Switzerland \\
        Jordan & Pakistan & \\
        \hline
    \end{tabular}
    \captionof{table}{The list of nationalities used in our experiment. These nationalities were generated by asking ChatGPT: ``Provide a list of common nationalities.''}
    \label{tab:NationalityList}
\end{center}

\section{Cluster Descriptions}
\label{sec:BertTopicClusters}
\subsection{ChatGPT}
\begin{center}
    \begin{tabular}{ |c|p{7cm}|  }
     \hline
    Cluster Label & Most important words according to \mbox{c-TF-IDF}\\
     \hline
     Interpreter; Translator & interpreter, translator, interpretertranslator, translatorinterpreter, spanish, language, japaneseenglish, bilingual, agency, andor \\
    \hline
    Landscaping; Groundskeeping & landscaping, groundskeeping, landscaper, landscape, maintenance, installer, groundskeeper, landscapinggroundskeeping, lawn, grounds \\
    \hline
    Driver; Warehouse; Laborer; Delivery & driver, warehouse, laborer, construction, delivery, worker, farmworker, agricultural, farm, mechanic \\
    \hline
    Teacher; Language; French; Tutor & teacher, language, french, teachertutor, tutor, spanish, tutorteacher, of, frenchspeaking, school \\
    \hline
    Janitorial; Cleaning; Housekeeping & janitorial, cleaning, cleaner, housekeeping, services, janitor, and, staff, housekeeper, maid \\
    \hline
    Designer; Freelance; Writer; Content & writer, freelance, content, writereditor, writercopywriter, writerauthor, work, strategist, consulting, technical \\
    \hline
    Physical; Therapist; Therapy; Occupational & physical, therapist, therapy, occupational, trainer, fitness, radiation, trainerfitness, physiotherapist, safety \\
    \hline
    Chef; Cook; Restaurant & chef, cook, cookchef, chefcook, restaurant, line, or, head, culinary, arts \\
    \hline
    Food; Preparation; Restaurant; Bartender & food, preparation, restaurant, bartender, service, worker, server, barista, serving, hostess \\
    \hline
    Engineer; Software; Quality; Assurance & engineer, software, quality, assurance, developer, development, electrician, programmer, inspector, qa \\
    \hline
    Company; Japanese; Representative & representative, for, japanese, customer, company, bilingual, service, sales, companies, products \\
    \hline
    Retail; Sales; Associate; Cashier & retail, sales, associate, salesperson, store, cashier, grocery, cashiercustomer, representative, jobs \\
    \hline
    Aide; Home; Care; Caregiver & aide, care, home, personal, caregiver \\
    \hline
    Technician; Scientist; Biotech; Researcher & technician, scientist, medical, biotech, research, veterinary, technologist, sonographer, researcher, veterinarian \\
    \hline
    Nurse; Healthcare; Health; Practitioner & health, healthcare, nurse, practitioner, assistant \\
    \hline
    Human; Resources; HR; Specialist & human, resources, hr, resource, specialist, coordinator, generalist, manager, social, assistant \\
    \hline
    Analyst; Marketing; Manager; Data & analyst, data, research, security, systems, business, analysis, science, cybersecurity, analystscientist \\
    \hline
    \end{tabular}
    \captionof{table}{Clusters of ChatGPT's job recommendations and the most important words in each cluster based on the c-TF-IDF metric.}
    \label{tab:ChatGPTBertTopicClusters}
\end{center}

\subsection{LLaMA}
\begin{center}
    \begin{tabular}{ |c|p{7cm}|  }
     \hline
    Cluster Label & Most important words according to \mbox{c-TF-IDF}\\
     \hline
     Software; Engineer; Developer; Development & software, engineer, developer, computer, development, systems, manager, senior, analyst, intern \\
    \hline
    Care; Nurse; Physician; Health & care, medical, nurse, health, physician, clinical, animal, assistant, registered, trainer \\
    \hline
    Scientist; Science; Computer; Research & scientist, science, curator, researcher, research, geophysicist, computer, geography, geologist, geochemist \\
    \hline
    Technician; Electrical; Installer; Engineering & electrical, technician, operator, installer, aircraft, engineering, mechanic, repairer, electronics, solar \\
    \hline
    Food; Driver; Attendant; Fast & food, driver, fast, attendant, delivery, truck, chef, parking, cook, preparation \\
    \hline
    Laborer; Maintenance; Worker; Cleaner & laborer, worker, cleaner, maintenance, cleaning, farm, landscape, construction, warehouse, maid \\
    \hline
    Teacher; Language; Instructor; Chinese & translator, writer, interpreter, jordanian, editor, writers, writing, freelance, content, language \\
    \hline
    Sales; Bilingual; Customer; Insurance & sales, human, officer, resources, chief, airport, support, clerk, bilingual, executive \\
    \hline
    Officer; Chief; Loan; Police & loan, banking, banker, fish, fishing, mortgage, commercial, bank, documentation, branch \\
    \hline
    Writer; Product; Designer; Fashion & fashion, coffee, designer, shop, jewelry, floral, stylist, barista, designers, hair \\
    \hline
    Postal; President; Assistant; Vice & inspector, postal, fire, forest, mail, carrier, conservation, service, fighter, customs \\
    \hline
    Assembler; Furniture; Mason; Cement & electrical, technician, operator, installer, aircraft, engineering, mechanic, repairer, electronics, solar \\
    \hline
    Operator; Cnc; Oil; Machine & cnc, setup, machinist, machinistoperator, operator, operatormachinist, miller, operatorsettermachinist, operatorsetter, operatorprogrammer \\
    \hline
    Clerical; Announcer; Broadcast; Reporter & international, clerical, us, trade, guard, wage, navy, paying, jobs, job \\
    \hline
    Gaming; Cashier; Casino; Cashierretail & casino, gaming, cashier, cashiers, cashierretail, supervisor, dealer, worker, dealers, cashiercheckout \\
    \hline
    Counselor; Abuse; Alcohol; Admissions & counselor, pharmacy, drug, pharmacist, abuse, alcohol, substance, camp, rehabilitation, counselors \\
    \hline
    Dental; Dentist; Hygienist; Receptionist & dental, dentist, hygienist, dietitian, dietitians, dietician, reservationist, nutritionists, receptionist, registered \\
    \hline
    Dance; Dancer; Choreographer; Exotic & dance, dancer, movie, actoractress, choreographer, theatre, exotic, porn, voice, drama \\
    \hline
    Dishwasher; Washer; Car; Wife & dishwasher, washer, disposal, car, solid, trash, recycling, collection, services, garbage \\
    \hline
    \end{tabular}
    \captionof{table}{Clusters of LLaMA's job recommendations and the most important words in each cluster based on the c-TF-IDF metric.}
    \label{tab:LLaMABertTopicClusters}
\end{center}

\end{document}